\documentclass[fleqn,10pt]{wlscirep}
\usepackage[utf8]{inputenc}
\usepackage[T1]{fontenc}
\usepackage[utf8]{inputenc}
\usepackage[per-mode=symbol]{siunitx} 
\usepackage{amsmath, amssymb}
\usepackage{mathtools}
\usepackage{physics}
\usepackage{hyperref}
\usepackage{xcolor}
\usepackage{tablefootnote}
\usepackage{multirow}
\usepackage{booktabs}
\definecolor{forestgreen}{RGB}{34 139 34}
\definecolor{darkorchid}{RGB}{153 50 204}
\definecolor{darkorange}{RGB}{255 140 0}

\title{Slack-based tunable damping leads to a trade-off between robustness and efficiency in legged locomotion}
\author[1,*]{An Mo}
\author[1,2]{Fabio Izzi}
\author[1]{Emre Cemal G\"onen}
\author[2,3]{Daniel Haeufle}
\author[1,4]{Alexander Badri-Spr\"owitz}
\affil[1]{Dynamic Locomotion Group, Max Planck Institute for Intelligent Systems, Stuttgart, 70569, Germany}
\affil[2]{Hertie Institute for Clinical Brain Research and Center for Integrative Neuroscience, University of T\"ubingen, T\"ubingen, 72076, Germany}
\affil[3]{Institute for Modelling and Simulation of Biomechanical Systems, Computational Biophysics and Biorobotics, University of Stuttgart, Stuttgart, 70569, Germany}
\affil[4]{Department of Mechanical Engineering, KU Leuven, Leuven, 3001, Belgium}

\affil[*]{mo@is.mpg.de}

\begin{abstract}

Animals run robustly in diverse terrain. This locomotion robustness is puzzling because axon conduction velocity is limited to a few ten meters per second. If reflex loops deliver sensory information with significant delays, one would expect a destabilizing effect on sensorimotor control. Hence, an alternative explanation describes a hierarchical structure of low-level adaptive mechanics and high-level sensorimotor control to help mitigate the effects of transmission delays.
Motivated by the concept of an adaptive mechanism triggering an immediate response, we developed a tunable physical damper system. Our mechanism combines a tendon with adjustable slackness connected to a physical damper. The slack damper allows adjustment of damping force, onset timing, effective stroke, and energy dissipation. We characterize the slack damper mechanism mounted to a legged robot controlled in open-loop mode. The robot hops vertically and planar over varying terrains and perturbations. 
During forward hopping, slack-based damping improves faster perturbation recovery (up to \SI{170}{\%}) at higher energetic cost (\SI{27}{\%}).
The tunable slack mechanism auto-engages the damper during perturbations, leading to a perturbation-trigger damping, improving robustness at minimum energetic cost. 
With the results from the slack damper mechanism, we propose a new functional interpretation of animals' redundant muscle tendons as tunable dampers.
\end{abstract}
\begin{document}

\flushbottom
\maketitle
\thispagestyle{empty}

% link to OneNote flowchart: \url{https://1drv.ms/u/s!AnCwl5jx0fzgg-okMrKLQlZUC3DuHA}\\
% link to Videos: \url{https://keeper.mpdl.mpg.de/d/755e320a76b3451a8e12/}\\
% link to data and CAD design: \url{https://keeper.mpdl.mpg.de/d/8fee69fa1fe7466b93bc/}

%!TEX root = main.tex
\section*{Introduction}

\begin{figure}[ht]
\centering
\includegraphics{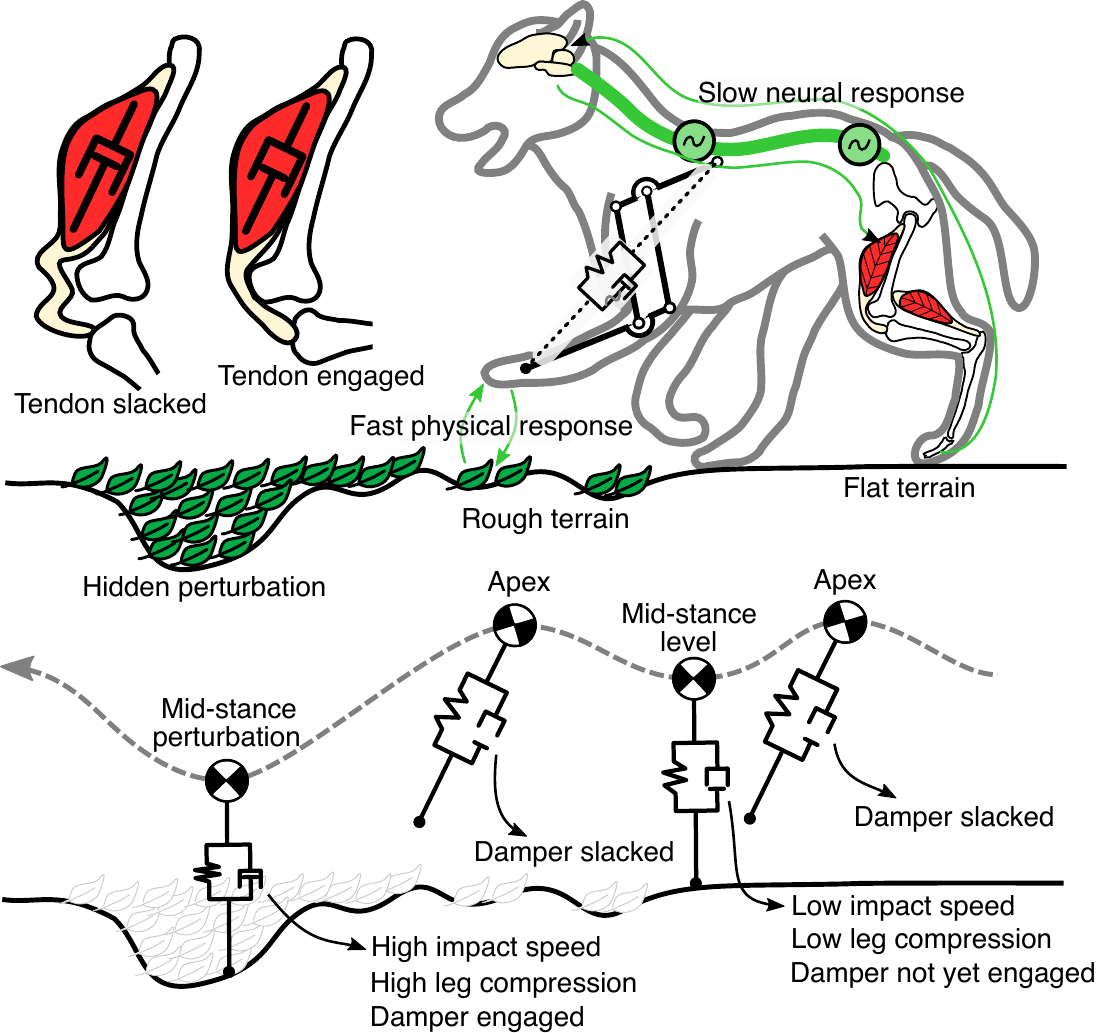}
\caption{\textbf{Top}: Fast running over ground perturbation is challenging. Due to sensorimotor delays up to \SI{50}{ms}, the central nervous system struggles to perceive and react to sudden ground disturbances \cite{more_scaling_2010}. In contrast, the intrinsic mechanics of the musculoskeletal system act like a spring damper. They produce a physical and, therefore, immediate (< \SI{5}{ms}) reaction when in contact with the environment. We hypothesize that the leg's damping mitigates ground disturbance through adaptive force production and energy dissipation. The tendon's slack, coupled with the joint's motion, auto-engages the damper. This creates a trade-off between locomotion robustness and energetic efficiency. \textbf{Bottom}: The damper slack enables perturbation-triggered damping. Sufficiently slacked, the damper does not engage during stance, and only spring-based torque is produced. When encountering a perturbation, the leg's compression increases further, removing all damper slack, and the damper engages in parallel to the spring.}
\label{fig:overview}
\end{figure}

% Excelling in fast evasive maneuvers over rough terrain is an evolutionary advantage for animals. 
Animals run dynamically over a wide range of terrain (Fig.\,\ref{fig:overview}).
The unevenness and changing compliance of natural terrain demand the capability for fast and dynamic adaptation to unexpected ground conditions.
However, animals' neurotransmission delays slow down sensorimotor information propagation~\cite{gordon2020}, rendering a neuronal response impossible for as much as \SI{5}{\%} to \SI{40}{\%} of the stance phase duration, depending on the animal size~\cite{more_scaling_2010}.
How animals are able to produce and maintain highly dynamic movements despite delayed sensorimotor information is, therefore, a central question in neuroscience and biorobotics~\cite{more_scaling_2010,more_scaling_2018,kamska20203d,ashtiani2021hybrid}.

Inherent mechanical properties of muscles facilitate the rejection of unexpected perturbations~\cite{loeb_hierarchical_1999,Wagner1999,grillner1972,Daley2009}. 
Muscular tissue possesses nonlinear elastic and viscous-like mechanical properties, which adapt the muscle force instantly to changes in the length or contraction velocity of the muscle-tendon fibers.
These mechanical properties enable the neuro-musculoskeletal system to react to external perturbations with zero delay, a capacity termed ``preflex''~\cite{Brown1995,Brown2000}.

Intrinsic elasticity and its role in legged locomotion have been studied extensively~\cite{alexander_role_1982,Alexander1991,Hof1990,Biewener2000,Robertson2014}. 
For instance, tendons, which behave like nonlinear serial springs, store and release mechanical energy during ground contact~\cite{alexander_role_1982} and improve shock tolerance~\cite{Roberts2010}. 
Inspired by this, parallel and series elastic actuators have successfully been implemented in the design of legged robots~\cite{sprowitz2013towards,Grizzle2009,hubicki2016atrias,zhao2022exploring}, demonstrating improved robustness at low control effort. 
In contrast, the functional role that damping plays in legged locomotion is less studied and understood.

Damping may produce a force outcome that is adaptive to the impact velocity.
This adaptive force output may enhance the effective force output during impacts~\cite{muller_kinetic_2014}, minimize control effort~\cite{Haeufle2014a}, stabilize motion~\cite{Seipel_2012,Secer2013,Abraham2015} and reject unexpected disturbances~\cite{Haeufle2010a,Kalveram2012}. 
% Moreover, damping may produce an adaptive force outcome to the impact velocity.
Nevertheless, damping is usually minimized in the design of (bio)robotic systems as it may lead to increased energy consumption.
Interestingly, vertebrates seem capable of tuning the damping produced by their muscle fibers~\cite{Guenther2010a}. 
This suggests that tunable damping may be a solution for regulating damping forces and dissipating energy depending on the terrain conditions.

Tunable damping in biorobotics can be implemented through control~\cite{monteleone2022,candan2020design}, i.e., virtual damping.
Virtual damping poses substantial design constraints. It requires precise velocity estimation, high-frequency control (>\SI{1}{kHz}), strong actuators to produce sufficient peak forces, and means to dissipate the resulting heat effectively~\cite{Seok2015,Hutter2012,Havoutis2013,Kalouche_2007,grimminger_2019}. 
Alternatively, physical dampers can be mounted to the robot's joints~\cite{Vanderborght_2013}.  
Tuning damping with a physical damper mounted to a legged robot proved challenging. 
Setting a higher damping rate resulted in the expected higher forces, but at reduced leg compression and effective damper stroke~\cite{mo2020effective}. 
Consequently, the dissipated energy indicated by the work loop area did not increase.
Additionally, fix-mounted physical dampers operate continuously and dissipate energy during unperturbed level running. 
Instead, physical tunable damping should ideally be triggered by the perturbation itself.
The damper should engage and self-adjust according to the presence and severity of the ground disturbance experienced during running.

The tendon slack observed in muscle-tendon units~\cite{robi2013physiology} and animal-inspired robots~\cite{badri2022birdbot} provided us with a design template for implementing tunable damping in a legged system (Fig.\,\ref{fig:overview}).
By disengaging the damper from its joint via controlled tendon slack, we expect to adjust the onset, timing, and amount of damper engagement.
Moreover, the tendon slack allows for a perturbation-trigger strategy (Fig.\,\ref{fig:overview}Bottom).
During steady-state running, for example, on flat terrain, the leg compresses without saturating the tendon slack.
Once an unperceived ground perturbation increases leg compression further, the tendon displacement will exceed the tendon's slack and start to auto-engage the damper.
This strategy enables adaptive force output triggered by ground perturbations.

% Conclusive paragraph
% aim 
We implemented and tested a bio-inspired, physical tunable damping strategy based on tendon slack in this work.
We aimed at producing perturbation-triggered damping and improving robustness against ground perturbations.
% approach
We evaluated this design concept on a robotic leg during vertical and forward hopping, both in steady-state and perturbed conditions.
% findings
Unlike earlier designs~\cite{mo2020effective}, our slack damper mechanism enabled straightforward adjustment of the damper engagement and energy dissipation.
We observed improved hopping robustness due to the adaptive characteristics of our physical damping design, whereas the energetic cost increases.
The perturbation-triggered capacity of our slack damper mechanism allows for a more favorable trade-off between robustness and efficiency.

\section*{Results}
% ************************************************************
We designed three experiments to study the proposed design with a hydraulic damper mounted to a robotic leg joint (Table \ref{tab:exp_design}).
We tested damper slack values of 10, 6, 3, and 0 \si{mm} for all conditions. These settings span from full slack (\SI{10}{mm}, minimum effective damping) to no slack (\SI{0}{mm}, maximum effective damping).
An open-loop controller produced the robot leg's locomotion pattern. Without feedback, ground perturbations were invisible to this high-level control (neural circuits), and perturbations could only be compensated by low-level mechanics in the form of a physical response.

We used the vertical hopping setup to investigate the vertical component of locomotion, allowing ground reaction force (GRF) measurement in all steps (Fig.\,\ref{fig:design}d).
We introduced step-down perturbation to evaluate the robustness of the system.
We used the forward hopping setup, which mounts the leg on a boom structure, to simulate more realistic locomotion dynamics (Fig.\,\ref{fig:design}e).
We analyzed forward hopping performance on rough terrain and robustness against ramp-up-step-down perturbation.

All data can be found in Supplementary Table S3-5.

\begin{table}
\centering
\begin{tabular}{lcccc} 
\toprule
\textbf{Experiment} & \textbf{Terrain}                   & \begin{tabular}[c]{@{}c@{}}\textbf{Perturbation }\\\textbf{height}\end{tabular} & \begin{tabular}[c]{@{}c@{}}\textbf{No. of }\\\textbf{perturbation steps}\end{tabular} & \begin{tabular}[c]{@{}c@{}}\textbf{No. of }\\\textbf{repetitions}\end{tabular}  \\ 
\midrule
\multirow{2}{*}{Vertical hopping}                                                                        & \multirow{2}{*}{step-down}         & 10\% LL               & \multirow{2}{*}{1}                                                                    & \multirow{2}{*}{10}                                                             \\
                                                                                                         &                                    & 15\% LL               &                                                                                       &                                                                                 \\ 
\hline
\multirow{3}{*}{Forward hopping}                                                                         & flat terrain                       & 0 mm                  & \multirow{3}{*}{15}                                                                   & \multirow{3}{*}{4}                                                              \\
                                                                                                         & rough terrain                      & ${\pm}$5 mm                 &                                                                                       &                                                                                 \\
                                                                                                         & rough terrain                      & ${\pm}$10 mm                &                                                                                       &                                                                                 \\ 
\hline
\multirow{2}{*}{Forward hopping}                                                                         & \multirow{2}{*}{ramp-up-step-down} & 15\% LL               & \multirow{2}{*}{1}                                                                    & \multirow{2}{*}{10}                                                             \\
                                                                                                         &                                    & 30\% LL               &                                                                                       &                                                                                 \\
\bottomrule
\end{tabular}
\caption{Experiment design, all experiments are repeated with damper slack values of 10, 6, 3, and 0 mm, from maximum slack to no slack.}
\label{tab:exp_design}
\end{table}

\subsection*{Vertical hopping with step-down perturbation}
\begin{figure}[ht]
\centering
\includegraphics{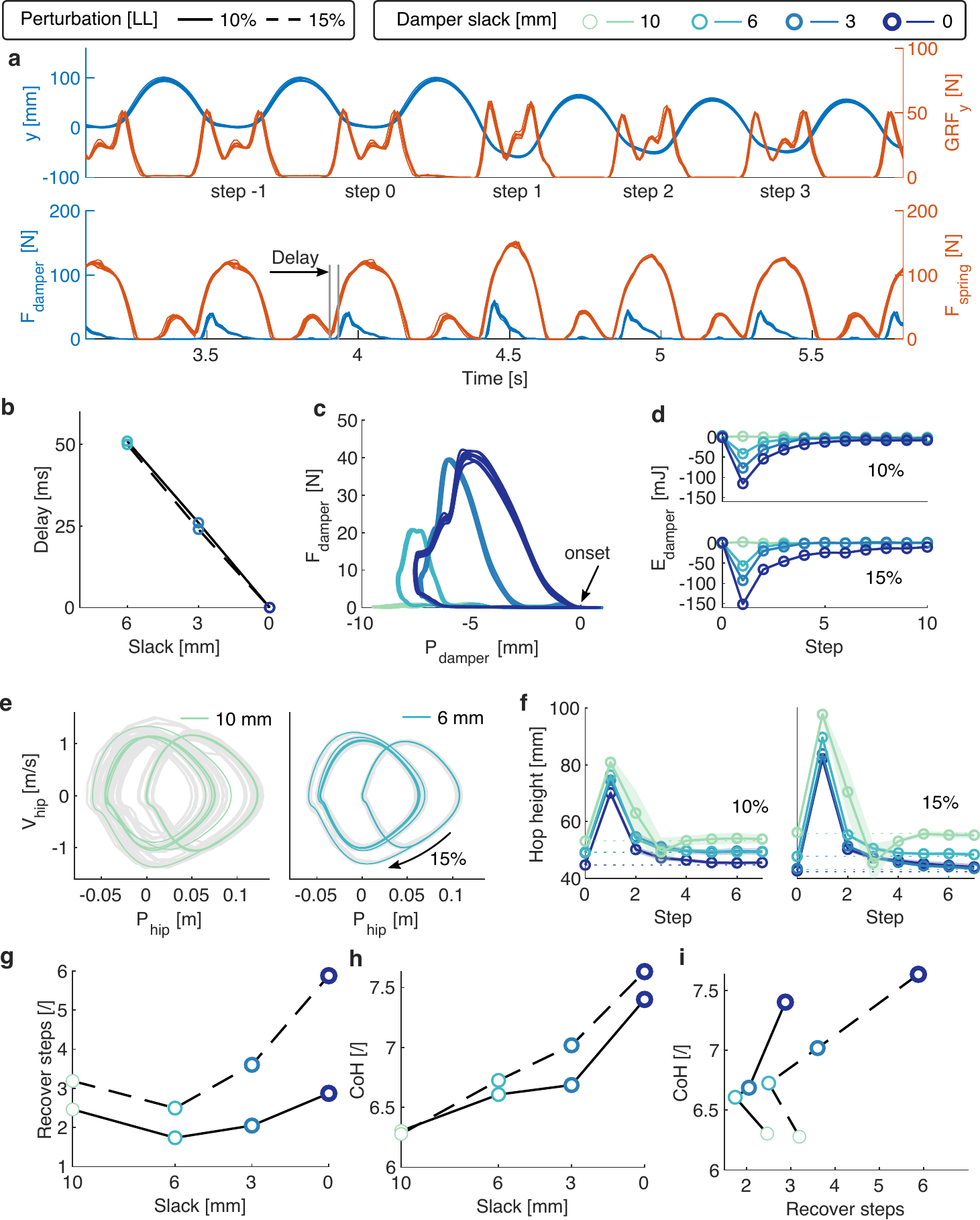}
\caption{Vertical hopping with step-down perturbation:  (\textbf{a}) 10-repetition-overlay time-series of hip position $y$, GRF, spring, and damper forces. 15\% LL perturbation at step 1 increases the GRF, spring and damper forces due to higher impact speed. The damper starts to produce force with a delay to touchdown due to the \SI{3}{mm} slack setting. (\textbf{b}) This damper engagement delay is adjustable by the damper slack setting. (\textbf{c}) The 10-repetition-overlay damper work loop in unperturbed periodic steps shows that the onset position can be reliably tuned and the standby dissipated energy (enclosed area) adjustable. (\textbf{d}) The average extra damper dissipated energy during perturbation steps. (\textbf{e}) Phase plot of hip position with \SI{10}{mm} and \SI{6}{mm} damper slack under 15\% LL perturbation. The grey overlay shows the overlap of 10 repetitions of 20 steps, while the darker line is the averaged trajectory. (\textbf{f}) The average hopping apex height during perturbation steps. The transparent overlay represents the 95\% confidence boundary. (\textbf{g}) The relationship between the number of steps to recovery after perturbation and the damper slack settings. (\textbf{h}) The relationship between the cost of hopping and the damper slack settings. (\textbf{i}) The relationship between the number of steps to recovery to the cost of hopping under different damper slack settings and perturbation levels.}
\label{fig:vertical}
\end{figure}

% ************************************************************
%hopping and perturbation recovery works
With feed-forward control, the leg hopped in the vertical setup for two perturbation levels and four slack values.
Figure \ref{fig:vertical}a shows an example of a time-series of 10 repetitions. 
The test condition included a perturbation of \SI{15}{\%} leg length (LL) and tendon slack of \SI{3}{mm} (Supplementary Movie S1).
At the perturbed step 1, the leg impacted the ground at a higher speed, compressing more. This resulted in higher damper and spring forces than during pre-perturbation levels.
We noticed that the damper force did not drop to zero at midstance due to the damper's internal recovery spring.

We found that the tunable slack mechanism was effective in tuning damping.
Damper slack adjustments of \SIrange{0}{6}{mm} resulted in a delayed engagement of the damper: from \SIrange{0}{50}{ms} after the onset of the spring force during level hopping (Fig.\,\ref{fig:vertical}b).
The damper's force-displacement work loops during level hopping confirmed the controllable onset of the damper force (Fig.\,\ref{fig:vertical}c). 
The enclosed work loop areas represent the damper's standby dissipated energy.
Damper slack values of 10, 6, 3, and \SI{0}{mm} can be mapped to standby dissipation of 152, 86, 29, and \SI{1}{mJ}.
At the perturbation step, the damper dissipated more energy (\SIrange{65}{190}{\%}) compared to level hopping standby dissipation (Fig.\,\ref{fig:vertical}d).
The extra dissipated energy is associated with the height of the ground drop, showing an adaptive energy dissipation to terrain disturbance.
In all tested conditions, the extra dissipated energy converged to 0 in the following steps, indicating recovery to steady-state hopping.

The robustness of the hopping system can be qualitatively assessed by the phase plot of the hip height (Fig.\,\ref{fig:vertical}e and Supplementary Movie S1).
With a \SI{10}{mm} slack setting, the hopping behavior was the most variable, as shown by the overlay of gray lines, representing 200 steps in 10 repetitions.
With a \SI{6}{mm} slack setting, the phase plot was clean, and the hopping converged to a new `limit cycle' in fewer steps than other settings.
A quantitative robustness measurement is the number of steps required to bring the system back to its original hopping height after the perturbation (Fig.\,\ref{fig:vertical}f).
The system's robustness was highest with the \SI{6}{mm} slack setting, requiring on average 1.7 and 2.5 steps to recover for \SI{10}{\%} and \SI{15}{\%} LL perturbation, respectively (Fig.\,\ref{fig:vertical}g).
At stronger perturbations, the robot needed more steps to recover.
% The recovery steps measurement shows a similar tendency as in the phase plots \alex{re-formulate, make sentence more precise}.
We measure the energetics of the hopping system by its cost of hopping (CoH, equation (\ref{eq:coh})).
The CoH increased from 6.3 to 7.6 with higher damping or stronger perturbations (Fig.\,\ref{fig:vertical}h).
With a damper slack of \SI{6}{mm} at \SI{10}{\%} LL perturbation, we found \SI{47}{\%} faster perturbation recovery in combination with \SI{5}{\%} higher CoH compared to \SI{10}{mm} damper slack (Fig.\,\ref{fig:vertical}i).

% ************************************************************
\subsection*{Forward hopping with continuous perturbation}

\begin{figure}[ht]
\centering
\includegraphics{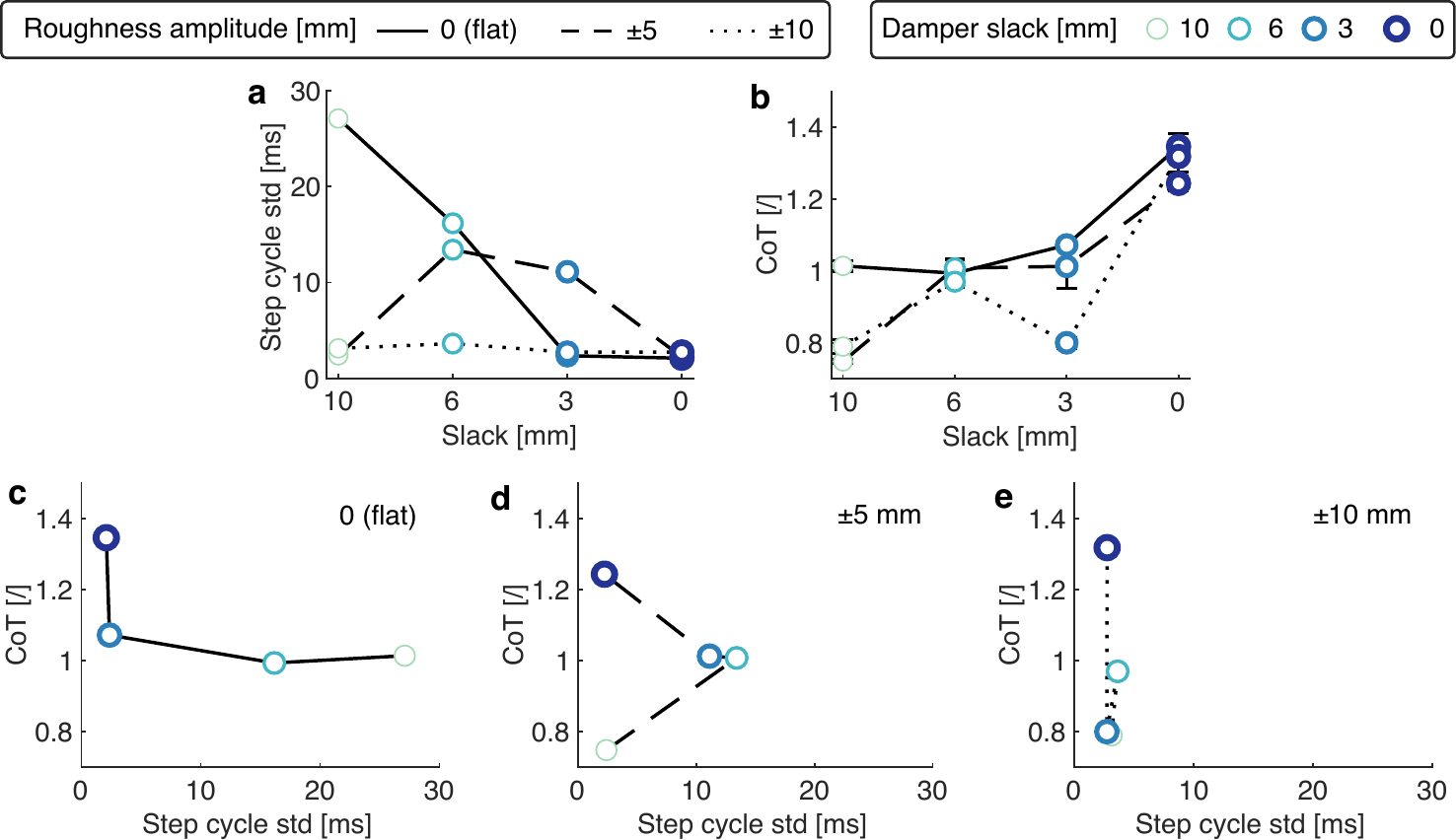}
\caption{Forward hopping with continuous perturbation: (\textbf{a}) The standard deviation of the step cycle time shows that the hopping periodicity is improved with higher damping (less slack). (\textbf{b}) The relationship between the CoT and the damper slack settings. (\textbf{c}) In flat terrain, the robot's ability to maintain periodic hopping is improved by higher damping at the cost of CoT. (\textbf{d, e}) In the continuous perturbation terrain, high damping is also associated with high CoT and good periodicity.}
\label{fig:continuous}
\end{figure}

During forward hopping on the sinusoidal ground, the standard deviation of the step cycle time quantifies the hopping periodicity.
In the flat terrain, the standard deviation of the step cycle time decreased from \SI{27}{ms} to \SI{2}{ms} with less damper slack, showing improved hopping periodicity with more damping (Fig.\,\ref{fig:continuous}a).
This tendency was less apparent in $\pm$5 and \SI{\pm 10}{mm} rough terrain, as the step cycle time variation increased first for the damper slack value \SI{6}{mm}, then decreased with less damper slack.
The energetic cost of forward hopping was measured as the cost of transport~\cite{tucker1975energetic} (CoT, equation (\ref{eq:cot})).
The CoT increased from a minimum of 0.75 to 1.35 with increasing damping (Fig.\,\ref{fig:continuous}b).
Both hopping periodicity and CoT were affected by the terrain's roughness.
In flat terrain, increasing damping was associated with improved periodicity and increased CoT (Fig.\,\ref{fig:continuous}c).
At $\pm$\SI{5}{mm} terrain roughness, data for damper slack values of 0, 3, and \SI{6}{mm} show similar tendency. The \SI{10}{mm} damper slack shows the best performance with a CoT of 0.75 and a standard deviation of \SI{2}{ms} cycle time (Fig.\,\ref{fig:continuous}d).
With $\pm$\SI{10}{mm} terrain roughness, the cycle time standard deviation was clustered around \SIrange{2}{3}{mm} for all slack settings, while the CoT varied from 0.79 to 1.32.
Among these three tested terrains, the strongest damping, i.e., the setting with a slack of \SI{0}{mm}, showed better periodicity with a cycle time standard deviation of $\approx$\SI{2}{ms}, but with the highest CoT, ranging from 1.24 to 1.35.

% ************************************************************
\subsection*{Forward hopping with ramp-up-step-down perturbation}

\begin{figure}[ht]
\centering
\includegraphics{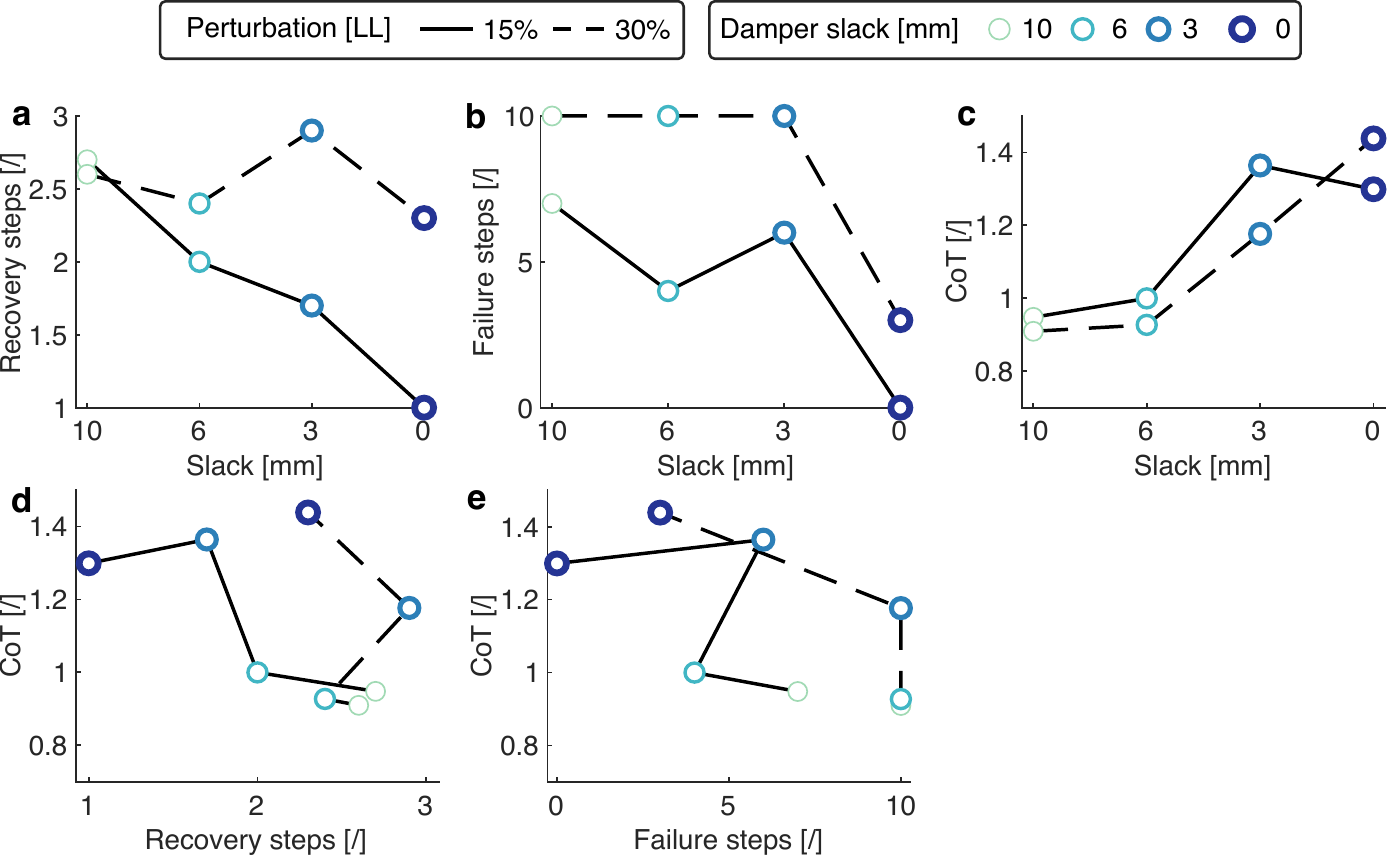}
\caption{Forward hopping with ramp-up and step-down perturbation: The robustness of the robot system is quantified with the number of steps needed to recover stable hopping (\textbf{a}) and the number of failed trials in 10 attempts (\textbf{b}). (\textbf{c}) The relationship between the CoT and the damper slack settings. (\textbf{d} and \textbf{e}) show the trade-off between robustness and CoT.}
\label{fig:step}
\end{figure}

We evaluated the system's robustness during forward hopping by testing its response to unexpected, sudden perturbations.
Thus, we analyzed the robotic leg's behavior with step-down perturbations in its hopping path.
As robustness measurement, we counted the number of steps required for the hopper to recover after the step perturbation.
The second measurement of robustness is the number of failures out of ten perturbation attempts.
By reducing the damper slack from \SI{10}{mm} to \SI{0}{mm}, the average recovery steps needed by the robotic leg decreased from 2.7 to 1.0 for the 15\% LL perturbation and from 2.6 to 2.3 for the \SI{30}{\%} LL perturbation (Fig.\,\ref{fig:step}a).
Similarly, with more damping, the number of failed trials decreased from 7 to 0 for the \SI{15}{\%} LL perturbation and 10 to 3 for the \SI{30}{\%} LL perturbation (Fig.\,\ref{fig:step}b).
The legged robot was less robust against a stronger perturbation, as it required on average 0.7 more recovery steps or failed, on average, four times more for the two tested perturbation levels.
Similar to the other two experiments, the energetic cost of the system increased with more damping, as the CoT increased from 0.95 to 1.44 (Fig.\,\ref{fig:step}c).
With a damper slack of \SI{0}{mm} at \SI{15}{\%} LL perturbation, we found \SI{170}{\%} faster perturbation recovery in combination with \SI{27}{\%} higher CoH compared to \SI{10}{mm} damper slack (Fig.\,\ref{fig:step}d).
With both measurements of robustness, we observed a tendency of increasing robustness at the expense of more energetic cost with higher damping settings (Fig.\,\ref{fig:step}d and e).

%!TEX root = main.tex
\section*{Discussion}

% message of the paragraph: The slack damper mechanism allows effective tunable damping.
The slack damper mechanism allows effective tunable damping.
This has three consequences:
First, depending on the slack setting, the damper produces an immediate or delayed response to ground impacts (Fig.\,\ref{fig:vertical}b).
Second, the onset of the damper stroke can be reliably set by the tendon slack (Fig.\,\ref{fig:vertical}c).
Third, the mechanical work generated by the damper is tunable, as shown by the change in the size of the enclosed work loops (Fig.\,\ref{fig:vertical}c).
Such a level of tunability of the damper response was not possible in our previous, more canonical approach of controlling the damping rate of the same damper model (implemented in a two-segment leg) via orifice adjustment~\cite{mo2020effective}.
In contrast, adjusting the slack of the damper tendon provides an effective strategy for tuning embedded damping in the robotic leg.
The slack in the damper tendon system allows the parallel spring to soften the damper impact within tens of milliseconds after the foot touchdown.
As a result, the damper produced less force and stroke than scenarios with less slack (Fig.\,\ref{fig:vertical}c).

% comment
In the steps following a sudden drop in ground height, the additional gravitational energy results in \SIrange{20}{30}{\%} higher touchdown speeds. 
The damper force and negative work increase accordingly, providing a beneficial mechanical reaction to compensate for the perturbation (Fig.\,\ref{fig:vertical}d).
Therefore, our damper implementation produces mechanical work in an adaptive manner that is consistent with the perturbation level and tunable by just one parameter; the damper tendon slack.

% message of this paragraph: Mechanical damping results in higher robustness: benefit of the approach
After testing the tunability of our damper tendon system, we characterized its effect on locomotion robustness during vertical and forward hopping, with and without perturbations.
In general, damping improves the robustness of our system.
In the vertical hopping experiments, adding a small amount of damping (\SI{6}{mm} slack) led to the fastest recovery from step perturbations (Fig.\,\ref{fig:vertical}e and g).
Above a certain amount of damping, the robotic leg appears to be over-damped, as shown by the damper dissipated energy (Fig.\,\ref{fig:vertical}d) and the hopping height over steps (Fig.\,\ref{fig:vertical}f).
% The convergence to pre-perturbation behavior is smooth but requires more steps.
In forward hopping experiments, more damping improved hopping periodicity (Fig.\,\ref{fig:continuous}a) and robustness (Fig.\,\ref{fig:step}a and b) without the emergence of an over-damping threshold.
% During our testing, the ramp-up-step-down perturbation, with a drop height of up to 30\% leg length, is challenging for legged robots~\cite{sprowitz2013towards}.
Our system performed well in this perturbed condition.
It overcame the perturbation 64 times out of 80 trials, despite using the simple feed-forward open-loop controller for forward hopping motion.
% The best performance was found with \SI{6}{mm} slack for 15\% leg length perturbation, during which only one step was sufficient for the leg to recover, and no failures occurred.
Although no electronic sensors are utilized to detect and respond to the perturbations, the passive compliance embedded in the leg acts as an intrinsic system of mechanical sensors and actuators, which detect and respond immediately to external disturbances.
The adaptive force output from damping plays a key role.
Simulation studies~\cite{Seipel_2012,Abraham2015} and muscle experiments~\cite{wilson2001horses} have revealed the stabilizing effect of damping in legged locomotion.
We offer a biorobotic understanding of damping in improving locomotion robustness.

%message of this paragraph: There is a trade-off between CoT and robustness due to damping.
The improved robustness introduced by the damper system comes at an energetic cost.
Higher damping settings (less slack) result in higher energy costs for all the experiments (Fig.\,\ref{fig:vertical}i, Fig.\,\ref{fig:continuous}b, and Fig.\,\ref{fig:step}c).
This occurs because the actuator needs to produce more power to compensate for the lost energy by damping (Fig.\,\ref{fig:vertical}c and d) and achieve a steady-state hopping behavior.
Tunable damping leads to a trade-off between the robustness and energy cost of the system (Fig.\,\ref{fig:step}d and e), despite a certain degree of nonlinearity.
This trade-off implies that hopping can be either robust against perturbations but with a penalty in energy consumption, or be energy efficient but vulnerable to disturbance. Adjusting tendon slack allows for selecting a suitable compromise depending on the terrain.

% comment
The benefit of damping for legged systems remains a debate in the field~\cite{Cham2007dynanic,Seipel_2012,Heim2020damping}.
Most research on legged locomotion focuses on optimizing a single aspect, such as robustness, stability, or energy consumption. 
On the contrary, evolution in biology is likely not a single-objective optimization process.
Instead, we argue that a more holistic perspective is required to understand the interaction among the many performance metrics characterizing legged locomotion.
This study's results highlight how damping is a key to balancing the trade-off between robustness and energy consumption.

% tendon slack allows perturbation-trigger strategy
The advantage of our slack damping mechanism concerning energy consumption is that it allows a perturbation-triggered strategy.
The damper tendon slack can be tuned to barely engage at level hopping. It will then engage once a ground perturbation induces higher impact velocities.
In this way, the absence of a damper minimizes the dissipating energy during level hopping, while the engagement of the damper improves robustness at ground perturbation steps.
This automatic on-off control was impossible with previous damper implementations~\cite{Garcia2011,Arelekatti2022}, because damping generated from friction, rheology, eddy currents, and fluid dynamics are hard to switch off completely~\cite{Vanderborght_2013}.
Instead of optimizing the adjustment of the nonlinear damping coefficient, our mechanism features a fixed damping coefficient but exploits a slack tendon to create a tunable on-off damping.
The proposed slack tendon could also be applied to selectively engage springs.
Hence, the tunable tendon slack mechanism offers a new mechanism for adaptive compliant actuator applications.

% future online tuning strategy
Besides the adaptive force output of damping, we expect the tunability of damping to provide more optimal hopping behavior, such as transitioning into new terrain.
When expecting a more uneven terrain, the damper slack can be adjusted accordingly to gain more robustness against the stronger perturbation.
This requires an online slack tuning mechanism and its feedback control strategy.
At the same time, we expect a feed-forward controller to be sufficient to produce highly robust running in an uncertain environment~\cite{wu20133}.
Limited by the hardware implementation, we did not thoroughly investigate an online tuning design.
Nevertheless, the four damper slack settings demonstrate the proof-of-concept of online tunable damping.

% The slack damping mechanism also allows adjusting the leg properties when transitioning into a new terrain.
% For example, when expecting a more uneven terrain, the damper slack can be adjusted accordingly to gain more robustness against the stronger perturbation.
% As such, a future design of a mechanical solution that modifies the slack level online might provide even more optimal hopping behavior.

% Conclusion
In summary, this work aims at understanding the tunable damping mechanism in legged locomotion.
% (core 1)
We proposed the slack damper strategy inspired by muscle tendon slack and tested it in robotic legged hopping.
The slack damper mechanism allows effective tunable damping regarding onset timing, engaged stroke, and energy dissipation.
% (core 2)
This study provides novel insights into the trade-off between energetics and robustness under different damping levels.
% (core 3)
Additionally, the slack damper design allows for perturbation-trigger damping, resolving the trade-off during locomotion with unexpected perturbation.
Our results could inspire future robotic locomotion hardware and controller design.

% missing in discussion
% system's natural dynamics
% limitation of the study

% % Take-home message
% - (core 1) Slack-damper system allows tunable damping [engineering novelty]
% - (core 2) Tunable damping allows for a trade-off between energetics and motion robustness [biomechanics finding]
% - (core 3) slack-design allows for perturbation triggered damping, making the trade-off more favourable [combine core 1 and core 2]
% - (future work) Task-dependent results suggest the need for more testing with more complex robotic system and real scenarios to optimze the use of physical damping. [optional]
%!TEX root = main.tex
\section*{Methods}

\subsection*{Biorobotic leg implementation}

The 3-segment leg design was inspired and simplified from the leg anatomy of small mammalian quadrupeds (Fig.\,\ref{fig:design}a).
It consisted of four links forming a pantograph structure (Fig.\,\ref{fig:design}b).
A spring and a damper coupled to the knee joint mimicked the passive compliance of the quadriceps muscles.
The gastrocnemius muscle and Achilles tendon were simplified as a rigid link to reduce parameter space.
The two-degrees-of-freedom leg was fully actuated by two motors (hip and knee).
The key design parameters are provided in the supplementary materials (Fig.\,S1 and Table S1).

The leg was fabricated mostly from off-the-shelf components and 3D-printing (Fig.\,\ref{fig:design}c).
The main structural components were 3D-printed using polylactic acid (PLA), except for the foot segment, which was 3D-printed using carbon-fiber-reinforced nylon to improve strength and impact resistance.
The hip and knee motors (MN7005-KV115, \textit{T-motor}, \SI{1.3}{Nm} maximum rated torque) were placed co-axially at the hip to reduce leg swing inertia, using a 5:1 planetary gearbox (lgu35-s, \textit{Matex}) to gear them down.
The knee torque was transmitted by a timing belt (SYNCHROFLEX 10/T5/390, \textit{Contitech}) with an additional 25:12 gear ratio.
We mounted two loadcells (model 3134, \textit{Phidgets}, \SI{20}{kg}) to the spring and the damper to measure forces.
The knee spring (SWS14.5-15, \textit{MISUMI}) was designed to hold the leg in stance.
Its stiffness of \SI{10.9}{N/mm} was empirically determined to generate three times the body weight of the robot at 10\% leg length deflection~\cite{Maarten1992,Walter2007}.
The knee damper (1210M, \textit{MISUMI}) was selected as the most effective damper from our previous study~\cite{mo2020effective}.
Both the spring and the damper were coupled to the knee joint through Dyneema tendons (Climax Combat Speed 250/150, \textit{Ockert}), with a cam radius of \SI{30}{mm} and \SI{20}{mm}, respectively.
A roller (VMRA20-4, \textit{MISUMI}) was attached to the piston of the damper to transform the tendon tension (``muscle lengthening'') in knee flexion to a push motion on the damper piston.
The whole leg weighs \SI{0.94}{kg}, with a resting leg length of \SI{31}{cm}.

\subsection*{Slack damper mechanism}
% Definition of design problem
Tuning an adjustable damper when operating within a legged system is challenging. 
Higher damping settings make the damper produce larger forces, which in turn can reduce the piston displacement, compromising the projected change in dissipated energy~\cite{mo2020effective}. 
Therefore, it is difficult to anticipate how adjusting the orifice of the damper internal valve affects the dissipated energy.
% Proposed solution - bioinspiration explanation
Instead of regulating the damper's force by adjusting the orifice size, we propose damping control by adjustment of the damper tendon slack.
Tendon slack has been observed in biology~\cite{robi2013physiology}, as tendons can stretch up to \SI{2}{\%} of their nominal length before starting to produce considerable force. This is called the \emph{toe-region} in the tendon's stress-strain diagram.
% Proposed solution - how it works
Inspired by this observation, we purposely set a defined length of slack in the tendon connecting the damper to the knee joint by using the thread connection between the damper and the loadcell (Fig.\,\ref{fig:design}c).
The design allowed for adjusting the tendon slack by turning the screw at the end of the piston ($\pm$\SI{1}{mm} per turn).

This damper slack mechanism permitted tunable damping because of two concomitant effects.
First, when the ground impact flexes the leg, the parallel spring decelerates leg flexion.
At the same time, the tendon slack saturates, thereby softening the engagement conditions for the damper's piston (less peak force).
Second, the tendon slack reduces the damper piston stroke (less displacement).
% At the same time, the tendon slack mitigates the impact condition for the damper, especially if a parallel spring is attached to the joint whose energy needs to be damped.
% How we implemented tendon slackness
% This extra time for the damper to engage allowed the knee stiffness (here realized by with a parallel spring) to mitigate the engagement condition for the damper (less peak force) due to the extra time decelerating the leg flexion.
% the knee stiffness (here realized by a parallel spring) combined with the extra time for the damper to engage, decelerates leg flexion while saturation of the tendon slack occurs, thereby softening the engagement conditions for the damper's piston (less peak force) and minimizing its travelling distance.
% The results of these two mechanisms, delayed and softened damper's engagement, is a more predictable energy dissipation by the damper.
The combination of these two mechanisms---softened (less peak force) and delayed (more slack $\hat{=}$ less displacement) damper engagement---makes the integrated damper energy dissipation more predictable.

\subsection*{Experimental setup}
We designed two experimental setups and three perturbation types to evaluate the proposed design in four slack settings.

The vertical hopping setup (Fig.\,\ref{fig:design}d) investigates only the vertical component of locomotion.
Such a reduced-order experiment reduced system complexity, allowing ground reaction force (GRF) measurement in all steps.
The forward hopping setup (Fig.\,\ref{fig:design}e) fixed the leg on a boom structure, simulating more realistic locomotion dynamics and allowing for more perturbation types.

We focus the investigation on the mechanical response produced by the passive damping embedded in the leg design.
Hence, we designed an open-loop locomotion controller such that it could not detect ground perturbation.
We tested three types of ground perturbations: step-down perturbation representing a sudden, unexpected disturbance during fast running;
continuous perturbation simulating rough terrain conditions, and
ramp-up-step-down perturbation combining gradual and sudden disturbance.

We tested damper tendon slack of 10, 6, 3, and \SI{0}{mm} for each test condition. The damper engaged synchronously with the spring in the \SI{0}{mm} slack setting. With the \SI{10}{mm} slack setting, the damper never engaged. Hence, we investigated a wide range of possible slack conditions, from complete to zero tendon slack.

\subsubsection*{Vertical hopping}

In the vertical hopping setup (Fig.\,\ref{fig:design}d), the hip of the robot leg was fixed to a vertical rail (SVR-28, \textit{MISUMI}).
A force sensor (K3D60a, \textit{ME}) was used to measure the ground reaction force during hopping.
The step-down perturbation was realized using a 3D-printed block (PLA) and an automatic block-removal device.
The block was placed on top of the force sensor to elevate the ground.
Magnets were inserted into the block and the top plate of the force sensor to prevent relative sliding during the leg impact.
The block-removal device was a lever arm actuated by a servo motor (1235M, \textit{Power HD}). 
The arm pushed away the block during the aerial phase of a hopping cycle (Supplementary Movie S1).
This automatic block-removal device was needed to remove the perturbation block within the aerial hopping phase reliably (\SI{200}{ms} in our experiments).

The vertical hopping setup was instrumented as follows.
The hip position was measured by a linear encoder (AS5311, \textit{AMS}).
The loadcells (spring and damper) and the ground reaction force sensor readings were amplified (9326, \textit{Burster}) and then recorded by a microcontroller (Due, \textit{Arduino}) with internal 12-bit ADC.
The motor position was measured by a 12-bit rotary encoder (AEAT8800-Q24, \textit{Broadcom}).
We used an open-source motor driver (Micro-Driver~\cite{grimminger_2019}) for motor control, current sensing, and encoder reading, which runs dual motor field-oriented control at \SI{10}{kHz}.
We monitored the motor driver current with a current sensor (ACS723T-AB, \textit{Allegro Microsystems}). 
A second microcontroller (Uno, \textit{Arduino}) was implemented to control the servo motor for automatic block removal.
A single-board computer (Raspberry Pi 4B) was used to centralize and synchronize all sensor readings and motor commands in \SI{1}{kHz}.

We implemented a Raibert-like~\cite{raibert1986legged} open-loop controller for vertical hopping.
The hip was position controlled with a PD controller to keep a vertical posture.
The knee was torque controlled to produce a defined torque at a given duty cycle, typically during the second half of the stance phase.
Motor commands are illustrated in the inserted plots in Fig.\,\ref{fig:design}d.
Control parameters for a stable hopping gait were found through manual tuning, resulting in a \SI{450}{ms} cycle time with \SI{100}{ms} knee motor push-off.
The knee torque was tuned for each setting of the damper tendon slack to maintain the same hopping heights across tested conditions (Supplementary Table S2).

We tested two perturbation levels: \SI{31}{mm} and \SI{47}{mm}, equivalent to \SI{10}{\%} and \SI{15}{\%} of the leg length, respectively.
For each hopping trial, the robot hopped for \SI{1}{min}. We analyzed ten steps before and after the perturbation.
Each hopping condition was repeated ten times.
We recorded in total 80 trials; two perturbations $\times$ four slack settings $\times$ ten repetitions.

\subsubsection*{Forward hopping}

\begin{figure}[ht]
\centering
\includegraphics{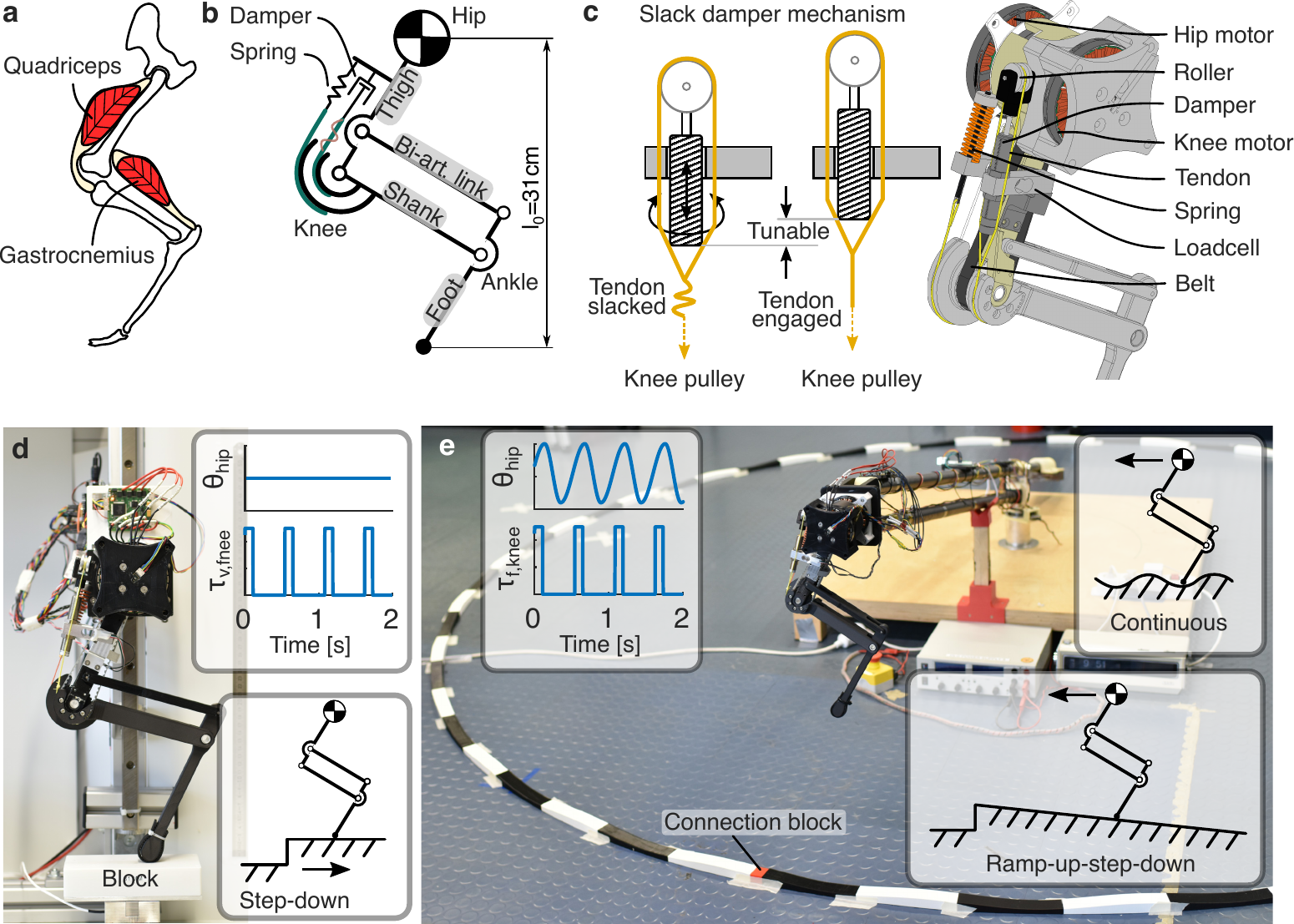}
\caption{Experiment setup overview. (\textbf{a}) Our leg design is inspired by the leg anatomy of mammalian quadrupeds. (\textbf{b}) We implement a pantograph leg design with spring and damper representing the passive compliance of the quadriceps and a biarticular segment simplifying the gastrocnemius muscle and the Achilles tendon. (\textbf{c}) The rendering of the design shows that the knee joint is coupled to the linear spring, linear damper through tendons, and the knee motor through a timing belt. By rotating the damper with a threaded connection to the loadcell, the damper will travel up and down, thus allowing space for tendon slack. (\textbf{d}) The vertical hopping setup fixes the robot leg on a vertical slider to test step-down perturbation, which is introduced by removing the perturbation block on top of the force sensor. The top right shows a feed-forward control pattern for hip position and knee torque. (\textbf{e}) The forward hopping setup fixes the robot leg on a rotary boom to test continuous perturbation (in photo) and ramp-up-step-down perturbation (Supplementary Movie S3). The top right shows a feed-forward CPG control pattern for hip position and knee torque.}
\label{fig:design}
\end{figure}

In the forward hopping setup (Fig.\,\ref{fig:design}e), the robot leg was mounted on a boom in a four-bar design. This mount permits only horizontal and vertical motion in the robot's sagittal plane.
The length of the boom was \SI{1.613}{m}, and the travel distance of a complete revolution was around \SI{10}{m}.
The boom design is openly available~\cite{boom2022}.

The instrumentation of the forward hopping setup was similar to that of the vertical hopping setup.
The force measurement and the automatic block-removal device were incompatible with the boom setup and were removed. All the other sensors remained.
Horizontal and vertical motions of the rotating boom were measured by two 11-bit rotary encoders (102-V, \textit{AMS}).

We generated the forward motion of the robot leg using a feed-forward central pattern generator (CPG).
In most vertebrates, CPGs contribute to controlling rhythmic motion~\cite{ijspeert2007swimming}, such as locomotion.
We implemented a CPG controller for the hip angle trajectory $\theta_{hip}$:
\begin{equation}\label{eq:hip}
\theta_{hip}=A_{hip} \cos(\Phi)+O_{hip}
\end{equation}
\begin{equation}\label{eq:phase}
\Phi= \begin{dcases}
\frac{\phi}{2D} & \phi<2\pi D_{vir}\\
\frac{\phi+2\pi(1-2D_{vir})}{2(1-D_{vir})} & \text{else}
\end{dcases}
% \left\{\begin{matrix}
% \frac{\phi}{2D}, \varphi<2\pi D_{vir}
% \\
% \frac{\phi+2\pi(1-2D_{vir}))}{2(1-D_{vir})}, else
% \end{matrix}\right.
\end{equation}
where $A_{hip}$ is the hip angle amplitude, $\Phi$ the hip angle phase, $O_{hip}$ the hip angle offset, $D_{vir}$ the virtual duty factor as the fraction of time when the leg moves forward, and $\phi$ the oscillator's linearly progressing phase.
The knee motor was torque controlled to generate push-off force in the late stance, following a fixed square-wave pattern as in the vertical hopping with the same frequency as the hip CPG.
The motor commands are shown in the overlay plots of Fig.\,\ref{fig:design}e.
For ease of comparison, the control parameters (Supplementary Table S2) remained the same for all forward-hopping experiments.

To replicate rough terrain in a controlled way, we designed 3D-print tracks with a sinusoidal profile (Fig.\,\ref{fig:design}e).
The circular track was built from 3D-printed blocks. These were serially connected and taped to the floor.
Each block is \SI{360}{mm} long, and 27 blocks fit the circumference of the hopping path.
A single, shorter connection block was added (red, Fig.\,\ref{fig:design}e).
This connection block prevents the hopping cycle from being entrained by the terrain harmonic perturbation pattern, e.g., repeatedly stepping onto the exact position of a cycle length of the track.
We tested two rough terrains, with the amplitude of the sinusoidal perturbation being \SI{5}{mm} and \SI{10}{mm}.
In addition, we also tested hopping on flat terrain.
For each trial, the robot performed a total of six revolutions.
We cropped the first and the last revolution from the recorded data and analyzed the remaining four revolutions (60 steps per condition).

Further, we designed ramp-up-step-down perturbations to disturb stable hopping during forward locomotion.
Within a revolution's \SI{10}{m} hopping path, we built a slope of \SI{3}{m} length for the robot leg to gradually climb and jump off.
We tested two perturbation heights: \SI{47}{mm} and \SI{93}{mm}, equivalent to 15\% and 30\% of leg length, respectively.
For each trial, the robot leg performed 12 revolutions.
We cropped the first and the last revolution from the recorded data and analyzed the remaining ten revolutions (150 steps per condition).

\subsubsection*{Data analysis}

The ground reaction force and vertical position data were filtered for the vertical hopping experiments with a 4th-order zero-lag Butterworth filter.
The loadcells were calibrated to output force reading only at leg flexion.
The spring and damper force data were smoothed using a moving average filter with a filter span of 5 samples.
The boom encoder data were filtered with a 4th-order zero-lag Butterworth filter for the forward hopping experiments.
The cutoff frequencies (\SI{9}{Hz}-\SI{19}{Hz}) of the Butterworth filter were determined by residual analysis~\cite{winter2009biomechanics}.

The recovery steps in the vertical hopping experiment were calculated by first computing the average hop height before perturbation as a reference height (dotted lines in Fig.\,\ref{fig:vertical}f) and then finding the post-perturbation hop height that intersected with the $\pm4\%$ boundary of the reference height~\cite{zhao2022exploring}.
The cost of hopping was calculated by normalizing the electric energy consumption $E_{elec}$ of one hopping step to the system's gravitational potential energy at the apex.
\begin{equation}\label{eq:coh}
CoH = \frac{E_{elec}}{m \cdot g \cdot h_{apex}}
\end{equation}
where $m$ is the robot mass, $g$ the gravitational acceleration, $h_{apex}$ the apex height position.

We defined two measurements for evaluating the robustness of forward hopping after the ramp-up-step-down perturbation.
The recovery steps were defined as the number of steps needed by the robot leg to recover its stable hopping after the step-down perturbation.
This metric quantified how fast the robot system can recover from perturbation, and it was measured by visual inspection of the video recordings and kinematic data.
The failure step metric quantified the number of failures after a perturbation was applied.
We identified two failure modes from the video recordings: the robot leg could slip or stop after the perturbation (Supplementary Movie S3).
The number of failures was visually counted from the video recordings.
The CoT was calculated by the electric energy consumption per distance traveled $d$, normalized by the robot weight.
\begin{equation}\label{eq:cot}
CoT = \frac{E_{elec}}{m \cdot g \cdot d}
\end{equation}

All data were processed with Matlab (R2021b, \textit{MathWorks}).

\section*{Data availability}
All data needed to evaluate the conclusions of the paper are available in the paper or the Supplementary Information. Data and scripts for data analysis and the computer-aided design model of the robot leg are available from \url{https://keeper.mpdl.mpg.de/d/8fee69fa1fe7466b93bc/}.
% this link will be replaced with Edmond repo later

\bibliography{literature}

\begin{thebibliography}{10}
\urlstyle{rm}
\expandafter\ifx\csname url\endcsname\relax
  \def\url#1{\texttt{#1}}\fi
\expandafter\ifx\csname urlprefix\endcsname\relax\def\urlprefix{URL }\fi
\expandafter\ifx\csname doiprefix\endcsname\relax\def\doiprefix{DOI: }\fi
\providecommand{\bibinfo}[2]{#2}
\providecommand{\eprint}[2][]{\url{#2}}

\bibitem{more_scaling_2010}
\bibinfo{author}{More, H.~L.} \emph{et~al.}
\newblock \bibinfo{journal}{\bibinfo{title}{Scaling of {Sensorimotor} {Control}
  in {Terrestrial} {Mammals}}}.
\newblock {\emph{\JournalTitle{Proceedings of the Royal Society B: Biological
  Sciences}}} \textbf{\bibinfo{volume}{277}}, \bibinfo{pages}{3563--3568},
  \doiprefix\url{10.1098/rspb.2010.0898} (\bibinfo{year}{2010}).

\bibitem{gordon2020}
\bibinfo{author}{Gordon, J.~C.}, \bibinfo{author}{Holt, N.~C.},
  \bibinfo{author}{Biewener, A.} \& \bibinfo{author}{Daley, M.~A.}
\newblock \bibinfo{journal}{\bibinfo{title}{Tuning of feedforward control
  enables stable muscle force-length dynamics after loss of autogenic
  proprioceptive feedback}}.
\newblock {\emph{\JournalTitle{eLife}}} \textbf{\bibinfo{volume}{9}},
  \bibinfo{pages}{e53908}, \doiprefix\url{10.7554/eLife.53908}
  (\bibinfo{year}{2020}).

\bibitem{more_scaling_2018}
\bibinfo{author}{More, H.~L.} \& \bibinfo{author}{Donelan, J.~M.}
\newblock \bibinfo{journal}{\bibinfo{title}{Scaling of sensorimotor delays in
  terrestrial mammals}}.
\newblock {\emph{\JournalTitle{Proc. R. Soc. B}}}
  \textbf{\bibinfo{volume}{285}}, \bibinfo{pages}{20180613},
  \doiprefix\url{10.1098/rspb.2018.0613} (\bibinfo{year}{2018}).

\bibitem{kamska20203d}
\bibinfo{author}{Kamska, V.}, \bibinfo{author}{Daley, M.} \&
  \bibinfo{author}{Badri-Spröwitz, A.}
\newblock \bibinfo{journal}{\bibinfo{title}{{3D} {Anatomy} of the {Quail}
  {Lumbosacral} {Spinal} {Canal}—{Implications} for {Putative}
  {Mechanosensory} {Function}}}.
\newblock {\emph{\JournalTitle{Integrative Organismal Biology}}}
  \textbf{\bibinfo{volume}{2}}, \doiprefix\url{10.1093/iob/obaa037}
  (\bibinfo{year}{2020}).

\bibitem{ashtiani2021hybrid}
\bibinfo{author}{Ashtiani, M.~S.}, \bibinfo{author}{Aghamaleki~Sarvestani, A.}
  \& \bibinfo{author}{Badri-Spr{\"o}witz, A.}
\newblock \bibinfo{journal}{\bibinfo{title}{Hybrid parallel compliance allows
  robots to operate with sensorimotor delays and low control frequencies}}.
\newblock {\emph{\JournalTitle{Frontiers in Robotics and AI}}}
  \bibinfo{pages}{170}, \doiprefix\url{10.3389/frobt.2021.645748}
  (\bibinfo{year}{2021}).

\bibitem{loeb_hierarchical_1999}
\bibinfo{author}{Loeb, G.~E.}, \bibinfo{author}{Brown, I.~E.} \&
  \bibinfo{author}{Cheng, E.~J.}
\newblock \bibinfo{journal}{\bibinfo{title}{A hierarchical foundation for
  models of sensorimotor control}}.
\newblock {\emph{\JournalTitle{Experimental Brain Research}}}
  \textbf{\bibinfo{volume}{126}}, \bibinfo{pages}{1--18},
  \doiprefix\url{10.1007/s002210050712} (\bibinfo{year}{1999}).

\bibitem{Wagner1999}
\bibinfo{author}{Wagner, H.} \& \bibinfo{author}{Blickhan, R.}
\newblock \bibinfo{journal}{\bibinfo{title}{Stabilizing function of skeletal
  muscles: an analytical investigation}}.
\newblock {\emph{\JournalTitle{Journal of Theoretical Biology}}}
  \textbf{\bibinfo{volume}{199}}, \bibinfo{pages}{163--179},
  \doiprefix\url{10.1006/jtbi.1999.0949} (\bibinfo{year}{1999}).

\bibitem{grillner1972}
\bibinfo{author}{Grillner, S.}
\newblock \bibinfo{journal}{\bibinfo{title}{The role of muscle stiffness in
  meeting the changing postural and locomotor requirements for force
  development by the ankle extensors}}.
\newblock {\emph{\JournalTitle{Acta Physiologica Scandinavica}}}
  \textbf{\bibinfo{volume}{86}}, \bibinfo{pages}{92--108},
  \doiprefix\url{10.1111/j.1748-1716.1972.tb00227.x} (\bibinfo{year}{1972}).

\bibitem{Daley2009}
\bibinfo{author}{Daley, M.~A.}, \bibinfo{author}{Voloshina, A.} \&
  \bibinfo{author}{Biewener, A.~A.}
\newblock \bibinfo{journal}{\bibinfo{title}{The role of intrinsic muscle
  mechanics in the neuromuscular control of stable running in the guinea
  fowl}}.
\newblock {\emph{\JournalTitle{The Journal of Physiology}}}
  \textbf{\bibinfo{volume}{587}}, \bibinfo{pages}{2693--2707},
  \doiprefix\url{10.1113/jphysiol.2009.171017} (\bibinfo{year}{2009}).

\bibitem{Brown1995}
\bibinfo{author}{Brown, I.~E.}, \bibinfo{author}{Scott, S.~H.} \&
  \bibinfo{author}{Loeb, G.~E.}
\newblock \bibinfo{journal}{\bibinfo{title}{{Preflexes---programmable high-gain
  zero-delay intrinsic responses of perturbed musculoskeletal systems}}}.
\newblock {\emph{\JournalTitle{Society of Neuroscience, Abstracts}}}
  \textbf{\bibinfo{volume}{21}}, \bibinfo{pages}{562} (\bibinfo{year}{1995}).

\bibitem{Brown2000}
\bibinfo{author}{Brown, I.~E.} \& \bibinfo{author}{Loeb, G.~E.}
\newblock \bibinfo{title}{A reductionist approach to creating and using
  neuromusculoskeletal models}.
\newblock In \emph{\bibinfo{booktitle}{Biomechanics and neural control of
  posture and movement}}, \bibinfo{pages}{148--163},
  \doiprefix\url{10.1007/978-1-4612-2104-3_10} (\bibinfo{publisher}{Springer},
  \bibinfo{year}{2000}).

\bibitem{alexander_role_1982}
\bibinfo{author}{Alexander, R.}, \bibinfo{author}{Maloiy, G.},
  \bibinfo{author}{Ker, R.}, \bibinfo{author}{Jayes, A.} \&
  \bibinfo{author}{Warui, C.}
\newblock \bibinfo{journal}{\bibinfo{title}{The role of tendon elasticity in
  the locomotion of the camel ({Camelus} dromedarius)}}.
\newblock {\emph{\JournalTitle{Journal of Zoology}}}
  \textbf{\bibinfo{volume}{198}}, \bibinfo{pages}{293--313},
  \doiprefix\url{10.1111/j.1469-7998.1982.tb02077.x} (\bibinfo{year}{1982}).

\bibitem{Alexander1991}
\bibinfo{author}{Alexander, R.~M.}
\newblock \bibinfo{journal}{\bibinfo{title}{{Energy-saving mechanisms in
  walking and running.}}}
\newblock {\emph{\JournalTitle{The Journal of Experimental Biology}}}
  \textbf{\bibinfo{volume}{160}}, \bibinfo{pages}{55--69},
  \doiprefix\url{10.1242/jeb.160.1.55} (\bibinfo{year}{1991}).

\bibitem{Hof1990}
\bibinfo{author}{Hof, A.~L.}
\newblock \bibinfo{title}{{Effects of Muscle Elasticity in Walking and
  Running}}.
\newblock In \emph{\bibinfo{booktitle}{Multiple Muscle Systems}},
  \bibinfo{pages}{591--607}, \doiprefix\url{10.1007/978-1-4613-9030-5_38}
  (\bibinfo{publisher}{Springer New York}, \bibinfo{address}{New York, NY},
  \bibinfo{year}{1990}).

\bibitem{Biewener2000}
\bibinfo{author}{Biewener, A.~A.} \& \bibinfo{author}{Roberts, T.~J.}
\newblock \bibinfo{journal}{\bibinfo{title}{{Muscle and tendon contributions to
  force, work, and elastic energy savings: a comparative perspective.}}}
\newblock {\emph{\JournalTitle{Exercise and sport sciences reviews}}}
  \textbf{\bibinfo{volume}{28}}, \bibinfo{pages}{99--107}
  (\bibinfo{year}{2000}).

\bibitem{Robertson2014}
\bibinfo{author}{Robertson, B.~D.} \& \bibinfo{author}{Sawicki, G.~S.}
\newblock \bibinfo{journal}{\bibinfo{title}{{Exploiting elasticity: Modeling
  the influence of neural control on mechanics and energetics of ankle
  muscle-tendons during human hopping}}}.
\newblock {\emph{\JournalTitle{Journal of Theoretical Biology}}}
  \textbf{\bibinfo{volume}{353}}, \bibinfo{pages}{121--132},
  \doiprefix\url{10.1016/j.jtbi.2014.03.010} (\bibinfo{year}{2014}).

\bibitem{Roberts2010}
\bibinfo{author}{Roberts, T.~J.} \& \bibinfo{author}{Azizi, E.}
\newblock \bibinfo{journal}{\bibinfo{title}{{The series-elastic shock absorber:
  tendons attenuate muscle power during eccentric actions.}}}
\newblock {\emph{\JournalTitle{Journal of applied physiology (Bethesda, Md. :
  1985)}}} \textbf{\bibinfo{volume}{109}}, \bibinfo{pages}{396--404},
  \doiprefix\url{10.1152/japplphysiol.01272.2009} (\bibinfo{year}{2010}).

\bibitem{sprowitz2013towards}
\bibinfo{author}{Spr{\"o}witz, A.} \emph{et~al.}
\newblock \bibinfo{journal}{\bibinfo{title}{Towards dynamic trot gait
  locomotion: Design, control, and experiments with cheetah-cub, a compliant
  quadruped robot}}.
\newblock {\emph{\JournalTitle{The International Journal of Robotics
  Research}}} \textbf{\bibinfo{volume}{32}}, \bibinfo{pages}{932--950},
  \doiprefix\url{10.1177/0278364913489205} (\bibinfo{year}{2013}).

\bibitem{Grizzle2009}
\bibinfo{author}{Grizzle, J.}, \bibinfo{author}{Hurst, J.},
  \bibinfo{author}{Morris, B.}, \bibinfo{author}{Park, H.-W.} \&
  \bibinfo{author}{Sreenath, K.}
\newblock \bibinfo{title}{{MABEL, a new robotic bipedal walker and runner}}.
\newblock In \emph{\bibinfo{booktitle}{2009 American Control Conference}},
  \bibinfo{pages}{2030--2036}, \doiprefix\url{10.1109/ACC.2009.5160550}
  (\bibinfo{publisher}{IEEE}, \bibinfo{year}{2009}).

\bibitem{hubicki2016atrias}
\bibinfo{author}{Hubicki, C.} \emph{et~al.}
\newblock \bibinfo{journal}{\bibinfo{title}{Atrias: Design and validation of a
  tether-free 3d-capable spring-mass bipedal robot}}.
\newblock {\emph{\JournalTitle{The International Journal of Robotics
  Research}}} \textbf{\bibinfo{volume}{35}}, \bibinfo{pages}{1497--1521},
  \doiprefix\url{10.1177/0278364916648388} (\bibinfo{year}{2016}).

\bibitem{zhao2022exploring}
\bibinfo{author}{Zhao, G.}, \bibinfo{author}{Mohseni, O.},
  \bibinfo{author}{Murcia, M.}, \bibinfo{author}{Seyfarth, A.} \&
  \bibinfo{author}{Sharbafi, M.~A.}
\newblock \bibinfo{journal}{\bibinfo{title}{Exploring the effects of serial and
  parallel elasticity on a hopping robot}}.
\newblock {\emph{\JournalTitle{Frontiers in Neurorobotics}}}
  \doiprefix\url{https://doi.org/10.3389/fnbot.2022.919830}
  (\bibinfo{year}{2022}).

\bibitem{muller_kinetic_2014}
\bibinfo{author}{M\"uller, R.}, \bibinfo{author}{Tschiesche, K.} \&
  \bibinfo{author}{Blickhan, R.}
\newblock \bibinfo{journal}{\bibinfo{title}{Kinetic and kinematic adjustments
  during perturbed walking across visible and camouflaged drops in ground
  level}}.
\newblock {\emph{\JournalTitle{Journal of Biomechanics}}}
  \textbf{\bibinfo{volume}{47}}, \bibinfo{pages}{2286--2291},
  \doiprefix\url{10.1016/j.jbiomech.2014.04.041} (\bibinfo{year}{2014}).

\bibitem{Haeufle2014a}
\bibinfo{author}{Haeufle, D. F.~B.}, \bibinfo{author}{G{\"{u}}nther, M.},
  \bibinfo{author}{Wunner, G.} \& \bibinfo{author}{Schmitt, S.}
\newblock \bibinfo{journal}{\bibinfo{title}{{Quantifying control effort of
  biological and technical movements: An information-entropy-based approach}}}.
\newblock {\emph{\JournalTitle{Physical Review E}}}
  \textbf{\bibinfo{volume}{89}}, \bibinfo{pages}{012716},
  \doiprefix\url{10.1103/PhysRevE.89.012716} (\bibinfo{year}{2014}).

\bibitem{Seipel_2012}
\bibinfo{author}{Shen, Z.} \& \bibinfo{author}{Seipel, J.}
\newblock \bibinfo{journal}{\bibinfo{title}{A fundamental mechanism of legged
  locomotion with hip torque and leg damping}}.
\newblock {\emph{\JournalTitle{Bioinspiration \& biomimetics}}}
  \textbf{\bibinfo{volume}{7}}, \bibinfo{pages}{046010},
  \doiprefix\url{10.1088/1748-3182/7/4/046010} (\bibinfo{year}{2012}).

\bibitem{Secer2013}
\bibinfo{author}{Secer, G.} \& \bibinfo{author}{Saranli, U.}
\newblock \bibinfo{title}{{Control of monopedal running through tunable
  damping}}.
\newblock In \emph{\bibinfo{booktitle}{2013 21st Signal Processing and
  Communications Applications Conference (SIU)}}, \bibinfo{pages}{1--4},
  \doiprefix\url{10.1109/SIU.2013.6531557} (\bibinfo{publisher}{IEEE},
  \bibinfo{year}{2013}).

\bibitem{Abraham2015}
\bibinfo{author}{Abraham, I.}, \bibinfo{author}{Shen, Z.} \&
  \bibinfo{author}{Seipel, J.}
\newblock \bibinfo{journal}{\bibinfo{title}{{A Nonlinear Leg Damping Model for
  the Prediction of Running Forces and Stability}}}.
\newblock {\emph{\JournalTitle{Journal of Computational and Nonlinear
  Dynamics}}} \textbf{\bibinfo{volume}{10}}, \doiprefix\url{10.1115/1.4028751}
  (\bibinfo{year}{2015}).

\bibitem{Haeufle2010a}
\bibinfo{author}{Haeufle, D. F.~B.}, \bibinfo{author}{Grimmer, S.} \&
  \bibinfo{author}{Seyfarth, A.}
\newblock \bibinfo{journal}{\bibinfo{title}{{The role of intrinsic muscle
  properties for stable hopping - stability is achieved by the force-velocity
  relation}}}.
\newblock {\emph{\JournalTitle{Bioinspiration {\&} Biomimetics}}}
  \textbf{\bibinfo{volume}{5}}, \bibinfo{pages}{016004},
  \doiprefix\url{10.1088/1748-3182/5/1/016004} (\bibinfo{year}{2010}).

\bibitem{Kalveram2012}
\bibinfo{author}{Kalveram, K.~T.}, \bibinfo{author}{Haeufle, D. F.~B.},
  \bibinfo{author}{Seyfarth, A.} \& \bibinfo{author}{Grimmer, S.}
\newblock \bibinfo{journal}{\bibinfo{title}{{Energy management that generates
  terrain following versus apex-preserving hopping in man and machine.}}}
\newblock {\emph{\JournalTitle{Biological Cybernetics}}}
  \textbf{\bibinfo{volume}{106}}, \bibinfo{pages}{1--13},
  \doiprefix\url{10.1007/s00422-012-0476-8} (\bibinfo{year}{2012}).

\bibitem{Guenther2010a}
\bibinfo{author}{G\"{u}nther, M.} \& \bibinfo{author}{Schmitt, S.}
\newblock \bibinfo{journal}{\bibinfo{title}{{A macroscopic ansatz to deduce the
  Hill relation.}}}
\newblock {\emph{\JournalTitle{Journal of theoretical biology}}}
  \textbf{\bibinfo{volume}{263}}, \bibinfo{pages}{407--18},
  \doiprefix\url{10.1016/j.jtbi.2009.12.027} (\bibinfo{year}{2010}).

\bibitem{monteleone2022}
\bibinfo{author}{Monteleone, S.}, \bibinfo{author}{Negrello, F.},
  \bibinfo{author}{Catalano, M.~G.}, \bibinfo{author}{Garabini, M.} \&
  \bibinfo{author}{Grioli, G.}
\newblock \bibinfo{journal}{\bibinfo{title}{Damping in compliant actuation: A
  review}}.
\newblock {\emph{\JournalTitle{IEEE Robotics \& Automation Magazine}}}
  \textbf{\bibinfo{volume}{29}}, \bibinfo{pages}{47--66},
  \doiprefix\url{10.1109/MRA.2021.3138388} (\bibinfo{year}{2022}).

\bibitem{candan2020design}
\bibinfo{author}{Candan, S.~{\c{S}}.}, \bibinfo{author}{Karag{\"o}z, O.~K.},
  \bibinfo{author}{Yaz{\i}c{\i}o{\u{g}}lu, Y.} \& \bibinfo{author}{Saranl{\i},
  U.}
\newblock \bibinfo{title}{Design of a parallel elastic hopper with a wrapping
  cam mechanism and template based virtually tunable damping control}.
\newblock In \emph{\bibinfo{booktitle}{Dynamic Systems and Control
  Conference}}, vol. \bibinfo{volume}{84270}, \bibinfo{pages}{V001T05A009},
  \doiprefix\url{10.1115/DSCC2020-3278} (\bibinfo{organization}{American
  Society of Mechanical Engineers}, \bibinfo{year}{2020}).

\bibitem{Seok2015}
\bibinfo{author}{Seok, S.} \emph{et~al.}
\newblock \bibinfo{journal}{\bibinfo{title}{Design principles for
  energy-efficient legged locomotion and implementation on the mit cheetah
  robot}}.
\newblock {\emph{\JournalTitle{IEEE/ASME Transactions on Mechatronics}}}
  \textbf{\bibinfo{volume}{20}}, \bibinfo{pages}{1117--1129},
  \doiprefix\url{10.1109/TMECH.2014.2339013} (\bibinfo{year}{2015}).

\bibitem{Hutter2012}
\bibinfo{author}{Hutter, M.} \emph{et~al.}
\newblock \bibinfo{title}{Starleth: A compliant quadrupedal robot for fast,
  efficient, and versatile locomotion}.
\newblock In \emph{\bibinfo{booktitle}{Adaptive Mobile Robotics}},
  \bibinfo{pages}{483--490}, \doiprefix\url{10.1142/9789814415958_0062}
  (\bibinfo{publisher}{World Scientific}, \bibinfo{year}{2012}).

\bibitem{Havoutis2013}
\bibinfo{author}{Havoutis, I.}, \bibinfo{author}{Semini, C.},
  \bibinfo{author}{Buchli, J.} \& \bibinfo{author}{Caldwell, D.~G.}
\newblock \bibinfo{title}{Quadrupedal trotting with active compliance}.
\newblock In \emph{\bibinfo{booktitle}{Mechatronics (ICM), 2013 IEEE
  International Conference on}}, \bibinfo{pages}{610--616},
  \doiprefix\url{10.1109/ICMECH.2013.6519112} (\bibinfo{publisher}{IEEE},
  \bibinfo{year}{2013}).

\bibitem{Kalouche_2007}
\bibinfo{author}{{Kalouche}, S.}
\newblock \bibinfo{title}{Goat: A legged robot with 3d agility and virtual
  compliance}.
\newblock In \emph{\bibinfo{booktitle}{2017 IEEE/RSJ International Conference
  on Intelligent Robots and Systems (IROS)}}, \bibinfo{pages}{4110--4117},
  \doiprefix\url{10.1109/IROS.2017.8206269} (\bibinfo{year}{2017}).

\bibitem{grimminger_2019}
\bibinfo{author}{Grimminger, F.} \emph{et~al.}
\newblock \bibinfo{journal}{\bibinfo{title}{An {Open} {Force}-{Controlled}
  {Modular} {Robot} {Architecture} for {Legged} {Locomotion} {Research}}}.
\newblock {\emph{\JournalTitle{The IEEE Robotics and Automation Letters}}}
  \doiprefix\url{10.1109/LRA.2020.2976639} (\bibinfo{year}{2020}).

\bibitem{Vanderborght_2013}
\bibinfo{author}{Vanderborght, B.} \emph{et~al.}
\newblock \bibinfo{journal}{\bibinfo{title}{Variable impedance actuators: A
  review}}.
\newblock {\emph{\JournalTitle{Robotics and autonomous systems}}}
  \textbf{\bibinfo{volume}{61}}, \bibinfo{pages}{1601--1614},
  \doiprefix\url{10.1016/j.robot.2013.06.009} (\bibinfo{year}{2013}).

\bibitem{mo2020effective}
\bibinfo{author}{Mo, A.}, \bibinfo{author}{Izzi, F.}, \bibinfo{author}{Haeufle,
  D.~F.} \& \bibinfo{author}{Badri-Spr{\"o}witz, A.}
\newblock \bibinfo{journal}{\bibinfo{title}{Effective viscous damping enables
  morphological computation in legged locomotion}}.
\newblock {\emph{\JournalTitle{Frontiers in Robotics and AI}}}
  \textbf{\bibinfo{volume}{7}}, \bibinfo{pages}{110},
  \doiprefix\url{10.3389/frobt.2020.00110} (\bibinfo{year}{2020}).

\bibitem{robi2013physiology}
\bibinfo{author}{Robi, K.}, \bibinfo{author}{Jakob, N.},
  \bibinfo{author}{Matevz, K.} \& \bibinfo{author}{Matjaz, V.}
\newblock \bibinfo{journal}{\bibinfo{title}{The physiology of sports injuries
  and repair processes}}.
\newblock {\emph{\JournalTitle{Current issues in sports and exercise
  medicine}}} \bibinfo{pages}{43--86}, \doiprefix\url{10.5772/54234}
  (\bibinfo{year}{2013}).

\bibitem{badri2022birdbot}
\bibinfo{author}{Badri-Spr{\"o}witz, A.},
  \bibinfo{author}{Aghamaleki~Sarvestani, A.}, \bibinfo{author}{Sitti, M.} \&
  \bibinfo{author}{Daley, M.~A.}
\newblock \bibinfo{journal}{\bibinfo{title}{Birdbot achieves energy-efficient
  gait with minimal control using avian-inspired leg clutching}}.
\newblock {\emph{\JournalTitle{Science Robotics}}}
  \textbf{\bibinfo{volume}{7}}, \bibinfo{pages}{eabg4055},
  \doiprefix\url{10.1126/scirobotics.abg4055} (\bibinfo{year}{2022}).

\bibitem{tucker1975energetic}
\bibinfo{author}{Tucker, V.~A.}
\newblock \bibinfo{journal}{\bibinfo{title}{The energetic cost of moving about:
  walking and running are extremely inefficient forms of locomotion. much
  greater efficiency is achieved by birds, fish—and bicyclists}}.
\newblock {\emph{\JournalTitle{American Scientist}}}
  \textbf{\bibinfo{volume}{63}}, \bibinfo{pages}{413--419}
  (\bibinfo{year}{1975}).

\bibitem{wilson2001horses}
\bibinfo{author}{Wilson, A.~M.}, \bibinfo{author}{McGuigan, M.~P.},
  \bibinfo{author}{Su, A.} \& \bibinfo{author}{Van~den Bogert, A.~J.}
\newblock \bibinfo{journal}{\bibinfo{title}{Horses damp the spring in their
  step}}.
\newblock {\emph{\JournalTitle{Nature}}} \textbf{\bibinfo{volume}{414}},
  \bibinfo{pages}{895--899} (\bibinfo{year}{2001}).

\bibitem{Cham2007dynanic}
\bibinfo{author}{Cham, J.~G.} \& \bibinfo{author}{Cutkosky, M.~R.}
\newblock \bibinfo{journal}{\bibinfo{title}{{Dynamic Stability of Open-Loop
  Hopping}}}.
\newblock {\emph{\JournalTitle{Journal of Dynamic Systems, Measurement, and
  Control}}} \textbf{\bibinfo{volume}{129}}, \bibinfo{pages}{275--284},
  \doiprefix\url{10.1115/1.2718237} (\bibinfo{year}{2006}).

\bibitem{Heim2020damping}
\bibinfo{author}{Heim, S.}, \bibinfo{author}{Millard, M.},
  \bibinfo{author}{Le~Mouel, C.} \& \bibinfo{author}{Badri-Spr\"owitz, A.}
\newblock \bibinfo{journal}{\bibinfo{title}{A little damping goes a long way: a
  simulation study of how damping influences task-level stability in running}}.
\newblock {\emph{\JournalTitle{Biology Letters}}}
  \textbf{\bibinfo{volume}{16}}, \bibinfo{pages}{20200467},
  \doiprefix\url{10.1098/rsbl.2020.0467} (\bibinfo{year}{2020}).

\bibitem{Garcia2011}
\bibinfo{author}{Garcia, E.}, \bibinfo{author}{Arevalo, J.~C.},
  \bibinfo{author}{Munoz, G.} \& \bibinfo{author}{Gonzalez-de Santos, P.}
\newblock \bibinfo{journal}{\bibinfo{title}{Combining series elastic actuation
  and magneto-rheological damping for the control of agile locomotion}}.
\newblock {\emph{\JournalTitle{Robotics and Autonomous Systems}}}
  \textbf{\bibinfo{volume}{59}}, \bibinfo{pages}{827--839},
  \doiprefix\url{10.1016/j.robot.2011.06.006} (\bibinfo{year}{2011}).

\bibitem{Arelekatti2022}
\bibinfo{author}{Arelekatti, V. N.~M.}, \bibinfo{author}{Petelina, N.~T.},
  \bibinfo{author}{Johnson, W.~B.}, \bibinfo{author}{Major, M.~J.} \&
  \bibinfo{author}{Winter~V, A.~G.}
\newblock \bibinfo{journal}{\bibinfo{title}{{Design of a Four-Bar Latch
  Mechanism and a Shear-Based Rotary Viscous Damper for Single-Axis Prosthetic
  Knees}}}.
\newblock {\emph{\JournalTitle{Journal of Mechanisms and Robotics}}}
  \textbf{\bibinfo{volume}{14}}, \doiprefix\url{10.1115/1.4052804}
  (\bibinfo{year}{2021}).

\bibitem{wu20133}
\bibinfo{author}{Wu, A.} \& \bibinfo{author}{Geyer, H.}
\newblock \bibinfo{journal}{\bibinfo{title}{The 3-d spring--mass model reveals
  a time-based deadbeat control for highly robust running and steering in
  uncertain environments}}.
\newblock {\emph{\JournalTitle{IEEE Transactions on Robotics}}}
  \textbf{\bibinfo{volume}{29}}, \bibinfo{pages}{1114--1124},
  \doiprefix\url{10.1109/TRO.2013.2263718} (\bibinfo{year}{2013}).

\bibitem{Maarten1992}
\bibinfo{author}{Bobbert, M.~F.}, \bibinfo{author}{Yeadon, M.~R.} \&
  \bibinfo{author}{Nigg, B.~M.}
\newblock \bibinfo{journal}{\bibinfo{title}{Mechanical analysis of the landing
  phase in heel-toe running}}.
\newblock {\emph{\JournalTitle{Journal of Biomechanics}}}
  \textbf{\bibinfo{volume}{25}}, \bibinfo{pages}{223--234},
  \doiprefix\url{10.1016/0021-9290(92)90022-S} (\bibinfo{year}{1992}).

\bibitem{Walter2007}
\bibinfo{author}{Walter, R.~M.} \& \bibinfo{author}{Carrier, D.~R.}
\newblock \bibinfo{journal}{\bibinfo{title}{{Ground forces applied by galloping
  dogs}}}.
\newblock {\emph{\JournalTitle{Journal of Experimental Biology}}}
  \textbf{\bibinfo{volume}{210}}, \bibinfo{pages}{208--216},
  \doiprefix\url{10.1242/jeb.02645} (\bibinfo{year}{2007}).

\bibitem{raibert1986legged}
\bibinfo{author}{Raibert, M.}
\newblock \emph{\bibinfo{title}{Legged {Robots} that {Balance}}}.
\newblock Artificial {Intelligence} {Series} (\bibinfo{publisher}{MIT Press},
  \bibinfo{year}{1986}).

\bibitem{boom2022}
\bibinfo{author}{Ruppert, F.}, \bibinfo{author}{Aghamaleki~Sarvestani, A.},
  \bibinfo{author}{Heim, S.}, \bibinfo{author}{Mo, A.} \&
  \bibinfo{author}{Badri-Spröwitz, A.}
\newblock \bibinfo{title}{{Instrumented boom, CAD data}},
  \doiprefix\url{10.17617/3.RSO2AG} (\bibinfo{year}{2022}).

\bibitem{ijspeert2007swimming}
\bibinfo{author}{Ijspeert, A.~J.}, \bibinfo{author}{Crespi, A.},
  \bibinfo{author}{Ryczko, D.} \& \bibinfo{author}{Cabelguen, J.-M.}
\newblock \bibinfo{journal}{\bibinfo{title}{From swimming to walking with a
  salamander robot driven by a spinal cord model}}.
\newblock {\emph{\JournalTitle{Science}}} \textbf{\bibinfo{volume}{315}},
  \bibinfo{pages}{1416--1420}, \doiprefix\url{10.1126/science.1138353}
  (\bibinfo{year}{2007}).

\bibitem{winter2009biomechanics}
\bibinfo{author}{Winter, D.}
\newblock \emph{\bibinfo{title}{Biomechanics and Motor Control of Human
  Movement}} (\bibinfo{publisher}{Wiley}, \bibinfo{year}{2009}).

\end{thebibliography}

\section*{Acknowledgements}
The authors thank the International Max Planck Research School for Intelligent Systems (IMPRS-IS) for supporting An Mo, Fabio Izzi, Emre Cemal G\"onen, and the China Scholarship Council (CSC) for supporting An Mo. 
The authors thank Felix Ruppert and Alborz Aghamaleki Sarvestani for assisting the robot development.
The authors also thank Prof. Syn Schmitt and Prof. Martin Giese for inspiring discussions on the project.

\section*{Author contributions}
A.M., F.I., D.H., and A.B.-S. conceptualized the project,
A.M., D.H., and A.B.-S. conceived the experiments 
A.M. designed and implemented the robot and experimental setups, 
A.M. and E.C.G. conducted experiments and analyzed the data,
all authors interpreted and discussed the data,
A.M., F.I. and E.C.G. prepared the manuscript,
all authors reviewed the manuscript.

\section*{Funding}
This work was funded by the Deutsche Forschungsgemeinschaft (DFG, German Research Foundation) - 449912641, HA 7170/3.

\section*{Additional information}
\textbf{Competing interests} The authors declare no competing interests.

\end{document}

\maketitle
\thispagestyle{empty}

\textbf{The PDF file includes:}\\
\\
\indent\indent\indent Figure S1\\
\indent\indent\indent Tables S1 to S5\\
\indent\indent\indent Legends for movies S1 to S3\\
\\
\indent\textbf{Other Supplementary Material for this manuscript includes the following:}\\
\\
\indent\indent\indent Movies S1 to S3\\
\indent\indent\indent Matlab scripts for data analysis: \url{https://keeper.mpdl.mpg.de/d/4b9bc3bb7dcb4cbb81c7/}\\
\indent\indent\indent Computer-aided design of the robot: \url{https://keeper.mpdl.mpg.de/d/309550dd51664c01b92b/}\\
\pagebreak

\section{Leg design parameters}
%%%%%%%%%%%%%%%%%%%%%%%%%%%%%%%%%%%%%%%%%%%%%%%%%%%%%%%%%%%%%%%%%%%%%%%%%%%%%%%%%%%%%%%%%%%%%%%%%%%%%%%%%%%
\begin{figure}[ht]
\centering
\includegraphics{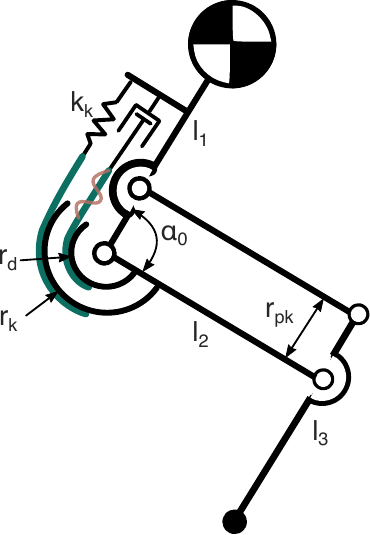}
\caption{Schematics of the leg with the key design parameters.}
\label{fig:design_para}
\end{figure}
%%%%%%%%%%%%%%%%%%%%%%%%%%%%%%%%%%%%%%%%%%%%%%%%%%%%%%%%%%%%%%%%%%%%%%%%%%%%%%%%%%%%%%%%%%%%%%%%%%%%%%%%%%%
\begin{table}[h]
\centering
\begin{tabular}{lcl}
\toprule
\textbf{Parameters} 						& 		& \textbf{Value} \\
\midrule
Robot mass - vertical hopping  	& $m_{v}$		& \SI{1.94}{kg} \\
Robot mass - forward hopping  	& $m_{f}$		& \SI{0.94}{kg} \\
Leg resting length		    	& $l_0$		& \SI{310}{mm} \\
Segment 1 length		    	& $l_1$		& \SI{150}{mm} \\
Segment 2 length		    	& $l_2$		& \SI{150}{mm} \\
Segment 3 length		    	& $l_3$		& \SI{150}{mm} \\
Knee spring pulley radius       & $r_k$		& \SI{30}{mm} \\
Knee damper pulley radius       & $r_d$		& \SI{20}{mm} \\
Knee spring stiffness 			& $k_k$		& \SI{10.9}{N/mm} \\
Bi-articular insertion radius	& $r_{pk}$   	& \SI{32}{mm} \\
Knee resting angle				& $\alpha_0$ 	& \ang{100} \\
\bottomrule
\end{tabular}
\caption{Robot design parameters.}
\label{tab:design_para}
\end{table}
%%%%%%%%%%%%%%%%%%%%%%%%%%%%%%%%%%%%%%%%%%%%%%%%%%%%%%%%%%%%%%%%%%%%%%%%%%%%%%%%%%%%%%%%%%%%%%%%%%%%%%%%%%%
\pagebreak
\section{Robot control parameters}
\begin{table}[h]
\centering
\begin{tabular}{lcl}
\toprule
\textbf{Parameters} 						& 									& \textbf{Value} \\
\midrule
\textbf{Vertical hopping} 						& 									&  \\
Hopping frequency       		& $f_{v}$          		& \SI{2.2}{Hz} \\
Knee torque amplitude       	& $\tau_{v}$          			& 4.0 - 4.3\si{Nm}  \\
Knee duty cycle	       			& $-$          			& 0.22  \\
\hline
\textbf{Forward hopping} 						& 									&  \\
Hip amplitude             		& $\theta_{hip}$          	& \ang{18}  \\
Hip offset             			& $O_{hip}$          		& \ang{2}  \\
Hopping frequency       			& $f_{f}$          			& \SI{1.85}{Hz}  \\
Hip virtual duty factor       	& $D_{vir}$          		& 0.4  \\
Knee torque amplitude       	& $\tau_{f}$          			& \SI{1.3}{Nm}  \\
Knee phase shift       			& $-$          			& 0.75 \\
Knee duty cycle	       			& $-$          			& 0.2  \\
\bottomrule
\end{tabular}
\caption{Robot control parameters}
\label{tab:control_para}
\end{table}
%%%%%%%%%%%%%%%%%%%%%%%%%%%%%%%%%%%%%%%%%%%%%%%%%%%%%%%%%%%%%%%%%%%%%%%%%%%%%%%%%%%%%%%%%%%%%%%%%%%%%%%%%%%
\pagebreak
\section{Experimental results}
% \todo{check abbr. for the figure}
\begin{table}[h]
\centering
\begin{tabular}{ccccccc} 
\toprule
\begin{tabular}[c]{@{}c@{}}\textbf{Perturbation }\\\textbf{[LL]}\end{tabular} & \begin{tabular}[c]{@{}c@{}}\textbf{Damper slack }\\\textbf{[mm]}\end{tabular} & \begin{tabular}[c]{@{}c@{}}\textbf{Hop height }\\\textbf{[mm]}\end{tabular} & \begin{tabular}[c]{@{}c@{}}\textbf{CoH }\\\textbf{[/]}\end{tabular} & \begin{tabular}[c]{@{}c@{}}\textbf{Recovery steps }\\\textbf{[/]}\end{tabular} & \begin{tabular}[c]{@{}c@{}}\textbf{Delay }\\\textbf{[ms]}\end{tabular} & \begin{tabular}[c]{@{}c@{}}$\textrm{\textbf{E}}_\textrm{d}$\\\textbf{[mJ]}\end{tabular}  \\ 
\midrule
10\%                                                                          & 10                                                                            & 53.3                                                                        & 6.3                                                                & 2.5                                                                            & -                                                                      & 1                                                                                   \\
10\%                                                                          & 6                                                                             & 49.3                                                                        & 6.6                                                                & 1.7                                                                            & 51                                                                     & 26                                                                                  \\
10\%                                                                          & 3                                                                             & 49.1                                                                        & 6.7                                                                & 2.0                                                                            & 26                                                                     & 117                                                                                 \\
10\%                                                                          & 0                                                                             & 44.7                                                                        & 7.4                                                                & 2.9                                                                            & 0                                                                      & 186                                                                                 \\
15\%                                                                          & 10                                                                            & 55.8                                                                        & 6.3                                                                & 3.2                                                                            & -                                                                      & 1                                                                                   \\
15\%                                                                          & 6                                                                             & 47.8                                                                        & 6.7                                                                & 2.5                                                                            & 50                                                                     & 29                                                                                  \\
15\%                                                                          & 3                                                                             & 43.2                                                                        & 7.0                                                                & 3.6                                                                            & 24                                                                     & 86                                                                                  \\
15\%                                                                          & 0                                                                             & 42.4                                                                        & 7.6                                                                & 5.9                                                                            & 0                                                                      & 152                                                                                 \\
\bottomrule
\end{tabular}
\caption{Experimental results of vertical hopping with step-down perturbation. The energy dissipated by the damper ($\textrm{E}_\textrm{d}$) is calculated by integrating the damping force with respect to the damper compression (Fig. 2c).}
\label{tab:exp_result_vert_hop}
\end{table}
%%%%%%%%%%%%%%%%%%%%%%%%%%%%%%%%%%%%%%%%%%%%%%%%%%%%%%%%%%%%%%%%%%%%%%%%%%%%%%%%%%%%%%%%%%%%%%%%%%%%%%%%%%%
\begin{table}[h]
\centering
\begin{tabular}{ccccc} 
\toprule
\begin{tabular}[c]{@{}c@{}}\textbf{Roughness amplitude}\\\textbf{[mm]}\end{tabular} & \begin{tabular}[c]{@{}c@{}}\textbf{Damper slack}\\\textbf{[mm]}\end{tabular} & \begin{tabular}[c]{@{}c@{}}\textbf{Speed }\\\textbf{[m/s]}\end{tabular} & \begin{tabular}[c]{@{}c@{}}\textbf{CoT }\\\textbf{[/]}\end{tabular} & \begin{tabular}[c]{@{}c@{}}\textbf{Step cycle std }\\\textbf{[ms]}\end{tabular}  \\ 
\midrule
0                                                                                   & 10                                                                           & 0.80                                                                    & 1.01                                                                & 27.1                                                                             \\
0                                                                                   & 6                                                                            & 0.79                                                                    & 0.99                                                                & 16.2                                                                             \\
0                                                                                   & 3                                                                            & 0.71                                                                    & 1.07                                                                & 2.4                                                                              \\
0                                                                                   & 0                                                                            & 0.67                                                                    & 1.35                                                                & 2.1                                                                              \\
$\pm$5                                                                                   & 10                                                                           & 0.76                                                                    & 0.75                                                                & 2.4                                                                              \\
$\pm$5                                                                                   & 6                                                                            & 0.76                                                                    & 1.01                                                                & 13.4                                                                             \\
$\pm$5                                                                                   & 3                                                                            & 0.74                                                                    & 1.01                                                                & 11.1                                                                             \\
$\pm$5                                                                                   & 0                                                                            & 0.68                                                                    & 1.24                                                                & 2.2                                                                              \\
$\pm$10                                                                                  & 10                                                                           & 0.76                                                                    & 0.79                                                                & 3.1                                                                              \\
$\pm$10                                                                                  & 6                                                                            & 0.71                                                                    & 0.97                                                                & 3.6                                                                              \\
$\pm$10                                                                                  & 3                                                                            & 0.72                                                                    & 0.80                                                                & 2.7                                                                              \\
$\pm$10                                                                                  & 0                                                                            & 0.66                                                                    & 1.32                                                                & 2.7                                                                              \\
\bottomrule
\end{tabular}
\caption{Experimental results of forward hopping with continuous perturbation.}
\label{tab:exp_result_cont_pert}
\end{table}
%%%%%%%%%%%%%%%%%%%%%%%%%%%%%%%%%%%%%%%%%%%%%%%%%%%%%%%%%%%%%%%%%%%%%%%%%%%%%%%%%%%%%%%%%%%%%%%%%%%%%%%%%%%
\begin{table}[h]
\centering
\begin{tabular}{cccccc} 
\toprule
\begin{tabular}[c]{@{}c@{}}\textbf{Perturbation}\\\textbf{[LL]}\end{tabular} & \begin{tabular}[c]{@{}c@{}}\textbf{Damper slack}\\\textbf{[mm]}\end{tabular} & \begin{tabular}[c]{@{}c@{}}\textbf{Speed}\\\textbf{[m/s]}\end{tabular} & \begin{tabular}[c]{@{}c@{}}\textbf{CoT}\\\textbf{[/]}\end{tabular} & \begin{tabular}[c]{@{}c@{}}\textbf{Recovery steps}\\\textbf{[/]}\end{tabular} & \begin{tabular}[c]{@{}c@{}}\textbf{Failure steps}\\\textbf{[/]}\end{tabular}  \\ 
\midrule
15\%                                                                         & 10                                                                           & 0.81                                                                   & 0.95                                                               & 2.7                                                                           & 7                                                                             \\
15\%                                                                         & 6                                                                            & 0.78                                                                   & 1.00                                                               & 2.0                                                                           & 4                                                                             \\
15\%                                                                         & 3                                                                            & 0.72                                                                   & 1.36                                                               & 1.7                                                                           & 6                                                                             \\
15\%                                                                         & 0                                                                            & 0.68                                                                   & 1.30                                                               & 1.0                                                                           & 0                                                                             \\
30\%                                                                         & 10                                                                           & 0.80                                                                   & 0.91                                                               & 2.6                                                                           & 10                                                                            \\
30\%                                                                         & 6                                                                            & 0.75                                                                   & 0.93                                                               & 2.4                                                                           & 10                                                                            \\
30\%                                                                         & 3                                                                            & 0.73                                                                   & 1.18                                                               & 2.9                                                                           & 10                                                                            \\
30\%                                                                         & 0                                                                            & 0.64                                                                   & 1.44                                                               & 2.3                                                                           & 3                                                                             \\
\bottomrule
\end{tabular}
\caption{Experimental results of forward hopping with ramp-up-step-down perturbation.}
\label{tab:exp_result_step_pert}
\end{table}
%%%%%%%%%%%%%%%%%%%%%%%%%%%%%%%%%%%%%%%%%%%%%%%%%%%%%%%%%%%%%%%%%%%%%%%%%%%%%%%%%%%%%%%%%%%%%%%%%%%%%%%%%%%
\pagebreak\section{Supplementary Videos}
\textbf{Movie-S1:} Vertical hopping with step-down perturbation. The leg is hopping on a block whose height is 15\% of the leg length. The slack of the damper is set to \SI{3}{mm}. The first part of the video shows the experiment in real-time. In the second part, slow motion of the same experiment is repeated. In both cases, hip position $y$, GRF, spring and damper forces are plotted synchronized to the video. In the last part, the phase plots of the all experiments show the relation between hopping speed and the hopping position.\\

\noindent\textbf{Movie-S2:} Forward hopping with continuous perturbation. The leg moves forward by hopping on the sinusoidal terrain with $\pm$\SI{10}{mm} amplitude. The damper is fully engaged, i.e., the slackness is \SI{0}{mm}. After the leg completes one full rotation on the terrain, the video shows frames taken by the high-speed video camera.\\

\noindent\textbf{Movie-S3:} Failure modes of forward hopping with ramp-up-step-down perturbation. The leg moves on the flat surface, and it gradually climbs on the ramp to jump off. The perturbation height is 30\% of the leg length, and the damper slack is set to \SI{3}{mm}. The bottom-right plot shows the synchronized hip position in planer motion. The video shows three cases: slipping, stopping, and the good response after the step-down perturbations. The slipping case can be identified by audio irregularity.

% \bibliography{literature}

% \begin{figure}[ht]
% \centering
% \includegraphics[width=\linewidth]{stream}
% \caption{Legend (350 words max). Example legend text.}
% \label{fig:stream}
% \end{figure}

% \begin{table}[ht]
% \centering
% \begin{tabular}{|l|l|l|}
% \hline
% Condition & n & p \\
% \hline
% A & 5 & 0.1 \\
% \hline
% B & 10 & 0.01 \\
% \hline
% \end{tabular}
% \caption{\label{tab:example}Legend (350 words max). Example legend text.}
% \end{table}

% Figures and tables can be referenced in LaTeX using the ref command, e.g. Figure \ref{fig:stream} and Table \ref{tab:example}.